\documentclass[11pt]{article}
\usepackage[preprint]{acl}
\usepackage[T1]{fontenc}
\usepackage{times,latexsym,microtype}
\usepackage{amsmath,amssymb,booktabs,graphicx,tabularx}
\usepackage{xurl,placeins}
\hypersetup{pdftitle={Collection-Date Assignment Redirects Answers in Temporal QA},pdfauthor={Minsu Oh, Seunghui Jwa, Soo Kyun Kim, Yeo-Chan Yoon}}
\renewcommand{\dbltopfraction}{0.9}
\title{Collection-Date Assignment Redirects Answers\\in Temporal QA}
\author{Minsu Oh$^{1}$ \quad Seunghui Jwa$^{1}$ \quad Soo Kyun Kim$^{2}$ \quad Yeo-Chan Yoon$^{1}$\thanks{Corresponding author.}\\[0.5ex]
\normalfont\normalsize $^{1}$Department of Artificial Intelligence, Jeju National University\\
\normalfont\normalsize $^{2}$Department of Computer Engineering, Jeju National University}
\begin{document}
\maketitle
\begin{abstract}
A collection date records when a source revision was acquired, not when
its contents changed. Can assigning this metadata redirect temporal
question answering despite unchanged evidence?
Swapping the question day and following day between two revisions lowers
later-reference exact match (EM) by 37.6--51.4 percentage points across
five open-weight LLMs on 595 Wikipedia pages, replicated on 128 disjoint pages.
Previous-day swaps retain 29.0--39.6-point losses on 497 pages.
Seven prompt variants, including explicit ignore-collection instructions
and archived-revision provenance, retain the effect. Fictional answers and
English month-name question dates show that remembered answers and
identical date strings are unnecessary.
Matched thinking reduces Qwen3.5's penalty from 30.1 to 11.1 points,
while native reasoners vary. Two API LLMs show small Wikipedia swap effects
and rarely shift to earlier answers on fictional records.
Large swap-induced losses concentrate around the question date;
swaps between distant dates have much smaller effects.
Adding fields containing one observed acquisition date lowers EM relative
to omission in three of five LLMs after correction.
The diagnostic isolates metadata effects from changing evidence or missing
the relevant revision, and provides a practical audit for temporal QA inputs.

\end{abstract}
\section{Introduction}

An older revision lists an athlete's former team; a newer revision lists
a different team (Table~\ref{tab:input}). Given both texts and revision timestamps,
can changing only collection dates redirect an answer about $t_1$
toward the earlier ($t_0$) reference answer?

Three times have different roles. The \emph{question date} is the time
asked about, not when a query is executed. A \emph{revision timestamp}
dates a source version. A \emph{collection date} records when that version
was acquired; collecting an older revision does not update its contents.
Acquisition metadata is available in real tool schemas: Diffbot exposes
\texttt{crawlTimestamp}, and Firecrawl's change-tracking response includes
\texttt{previousScrapeAt} \citep{diffbot2026dates,firecrawl2026scrape}.
These establish metadata availability, not how often it reaches LLM prompts.
Our question-day/previous-day comparison models an old mirror fetched today alongside
a newer canonical version cached yesterday: collection order need not
follow revision order.
VersionRAG and ContextNest manage revision identity and eligibility before
an LLM receives evidence \citep{huwiler2025versionrag,sulpovar2026contextnest};
we study the remaining answer-selection problem
after both revisions enter the context. Publication dates in FreshPrompt
\citep{vu2024freshllms} have a different temporal role.

\begin{figure*}[t]
\centering\includegraphics[width=0.90\textwidth]{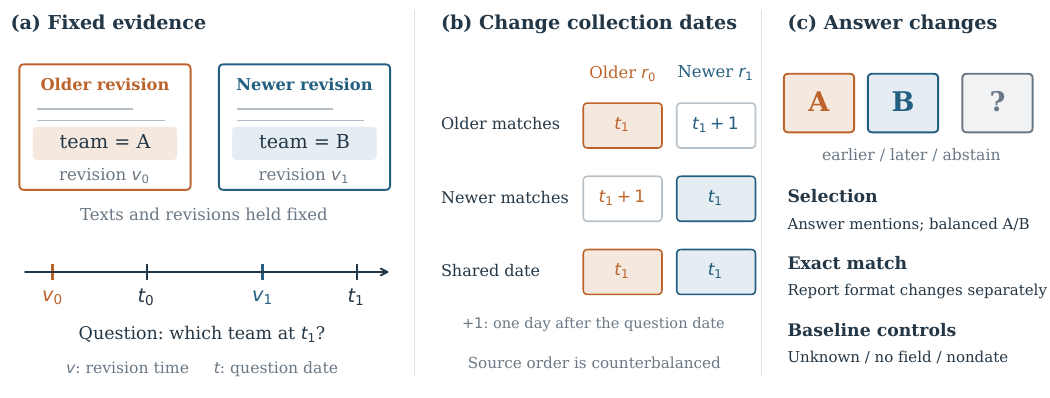}
\caption{Evidence-fixed collection-date interventions.
(a) Texts, revision timestamps and the question remain fixed.
(b) For a question about $t_1$, collection dates $r_0,r_1$ can match $t_1$ for the older
revision, the newer revision, or both; $+1$ means one day later.
(c) Exact match and answer mentions separate selection from formatting;
A/B answer selection probes shared dates, not swaps.
Missing-value, absent-field and nondate controls
test alternative explanations. The schematic is illustrative, not a result plot.}
\label{fig:overview}
\end{figure*}

We vary whether collection dates match or lie near the fixed question date
(Figure~\ref{fig:overview}). Equal dates are one condition, not the proposed
general cause. As an informative artificial positive control, we set each
revision's collection date to its associated question date. Unknown, absent-field and
nondate controls separate the effects of date values from prompt format.
We isolate metadata effects after retrieval; we do not aim to improve retrieval.

\begin{figure*}[t]
\centering\includegraphics[width=0.96\textwidth]{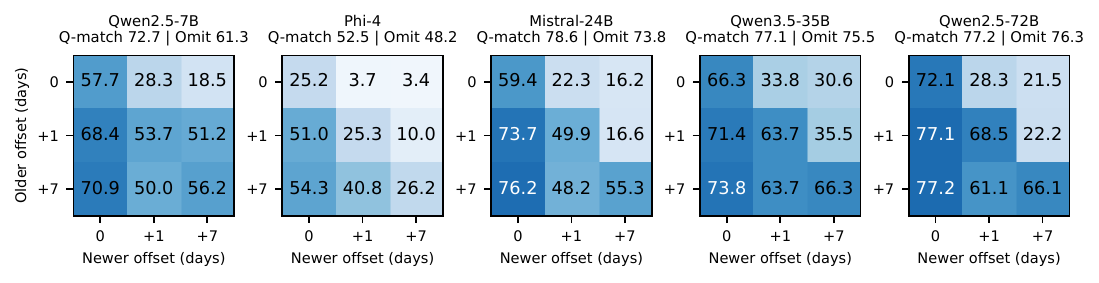}
\caption{Later-reference EM ($t_1$, \%) on 595 Wikipedia pages. Older/newer
revision collection offsets are days after question date $t_1$, averaged
over both source orders. Compare $(0,1)$ with $(1,0)$ for question-date
matching and $(1,7)$ with $(7,1)$ for proximity without an exact date match.
Texts and revision timestamps
remain fixed. Q-match supplies an informative positive control; omit
removes collection metadata.}
\label{fig:w2-factorial}
\end{figure*}

Swapping the same two collection-date values changes later-reference EM
($t_1$) in all five LLMs (Figure~\ref{fig:w2-factorial}). This result replicates
on nonoverlapping pages previously used for other conditions. Large accuracy
drops occur when dates match or are close to $t_1$; they do not establish a
general rule of selecting whichever date is nearest.
Sensitivity also occurs with administrative dates.

Three questions organize the analysis.
\textbf{RQ1:} Which collection-date assignments redirect answer choice
despite unchanged revision evidence?
\textbf{RQ2:} Do date direction, prompt controls and answer content
explain or remove this pattern?
\textbf{RQ3:} How does sensitivity vary with reasoning settings,
closed-book agreement and revision selection?

Our contribution is a controlled diagnosis of date assignment with revision
texts and timestamps fixed, replication on nonoverlapping page sets, and
controls for date ordering, instructions, wording and archive provenance.
Answer-content measures distinguish answer selection from changes in wording.
Reasoning and closed-book comparisons characterize changes in sensitivity.

\section{Task, Evidence and Measurements}
\label{sec:design}\label{sec:data}

\paragraph{Three times, two revisions.}
A question asks about date $t$. Each supplied revision has text $s$,
revision timestamp $v$, and collection date $r\geq v$.
Since $v_0\leq t_0<v_1\leq t_1$, only the older revision,
timestamped $v_0$, is available by the earlier question time $t_0$.
Both revisions are available by the later question time $t_1$;
the newer one, timestamped $v_1$, is then the latest.
Fetching an archived revision later does not update its text.
\begin{table*}[t]
\centering\footnotesize
\begin{tabularx}{\textwidth}{@{}lXX@{}}
\toprule
Question & \multicolumn{2}{l}{Which team does Jordan Harris (ice hockey) play for?} \\
\midrule
 & Older revision & Newer revision \\
Native revision & 2024-08-30T01:23:19Z & 2025-07-08T21:59:15Z \\
Recorded team & Columbus Blue Jackets & Boston Bruins \\
Question date ($t_0,t_1$) & 2024-09-01 & 2025-07-16 \\
Collection date: question-matched & 2024-09-01 ($r_0=t_0$) & 2025-07-16 ($r_1=t_1$) \\
Collection date: shared question date & 2025-07-16 ($r_0=t_1$) & 2025-07-16 ($r_1=t_1$) \\
Matching swap: older receives $t_1$ & 2025-07-16 ($r_0=t_1$) & 2025-07-17 ($r_1=t_1+1$) \\
Matching swap: newer receives $t_1$ & 2025-07-17 ($r_0=t_1+1$) & 2025-07-16 ($r_1=t_1$) \\
\bottomrule
\end{tabularx}
\caption{Paired-revision inputs. Collection-date rows are assigned controls,
not observed logs. The last two rows exchange the same dates between
revisions, the headline manipulation. Question, team text and revision
timestamps stay fixed. This is the original 595-page
cohort's shortest team-field pair by combined retrieved length,
selected without LLM outputs.
Official \href{https://www.nhl.com/canadiens/news/canadiens-acquire-patrik-laine-and-a-2026-second-round-pick}{August 2024}
and \href{https://www.nhl.com/bruins/news/bruins-announce-july-1-transactions-x3998}{July 2025}
transaction notices corroborate the team changes; they are not extra model inputs.
Experiments use both complete snippets, with both source orders.}
\label{tab:input}
\end{table*}

\paragraph{Sources and cohorts.}
DynaQuest \citep{lin2025dynaquest} supplies dated questions and reference
answers: the expected answers at each date. For example,
a player's team may be Team A at $t_0$ and Team B at $t_1$.
Availability and chronology checks retain 595, 596 and 599 distinct-page
cohorts from three separately selected 600-page samples, without replacement
(Appendix~\ref{app:data}). To construct LLM evidence inputs, BM25
\citep{robertson2009bm25} selects up to four 1,000-character passages
per revision using only the question, not its reference answers.
Answer presence is not required.
The 595-page cohort supports the main date-assignment experiment;
497 of those pages also meet the timestamp constraints for the
question-day/previous-day swaps below. Both use raw Wikipedia markup.
The separate 596-page cohort supports baseline controls and the
128-page swap replication. The 599-page cohort supports shared-date
and reasoning comparisons. These studies use plain text preserving
words and numbers. Table~\ref{tab:cohort-map} lists sample overlaps.
Within-study date contrasts fix the text; cross-study accuracy also
reflects cohort and rendering differences.

\paragraph{LLMs and outcomes.}
The five LLMs are Qwen2.5-7B/72B, Phi-4,
Mistral-Small-24B-2501 (Mistral-24B), and Qwen3.5-35B-A3B
(Qwen3.5-35B)
\citep{qwen2024model,microsoft2024phi,mistral2025small,qwen2026threefive}.
They use greedy short-answer generation with Qwen3.5's thinking disabled.
EM measures exact match with normalized reference aliases: accepted
answer forms (e.g., a benchmark may accept ``Manchester United'' and
``Manchester United FC'').
We also measure earlier-reference ($t_0$) EM, exclusive earlier/later answer inclusion, token
F1 and literal abstention. Alias inclusion detects answer content beyond
exact wording, but does not establish entailment.
Both aliases occur in the input excerpts on 448/596 pages; all 148 others remain included.
Appendix~\ref{app:statistics} details scoring and normalization.

\section{Controlled Date Assignments}
\label{sec:mechanism}

\paragraph{Assignment swaps.}
For questions about $t_1$, $(o,n)\in\{0,1,7\}^2$ gives older/newer
collection-date offsets from $t_1$, independent of display order.
The \emph{question-date matching swap}
compares $(0,1)$ with $(1,0)$. For a January 10 question, it exchanges
January 10 and 11 between the two revisions. The \emph{proximity swap}
compares $(1,7)$ with $(7,1)$: it instead exchanges January 11 and 17,
changing which revision's collection date is closer without an exact date match.
Both fix the date values, question, texts and revision timestamps.
To account for presentation order, we test older-first and newer-first
revision inputs for each assignment, averaging scores per page.
Since $v_0<v_1\leq t_1$, the newer revision's timestamp $v_1$ is on or
before $t_1$. A later acquisition does not imply a later text change.
The benchmark's reference answer at $t_1$ corresponds to that newer
revision; recovering it does not guarantee historical truth.
Archive provenance and ignore-collection controls test whether an
ambiguous live-source interpretation explains the losses.
Matching at $t_0$ would require the newer revision to be
collected at $t_0<v_1$, violating collection after revision. We instead
test proximity with both collection dates after both revisions and
questions, on the same 128 pages at both question times
(Appendix~\ref{app:remaining-local}).

\paragraph{Past-side dates and prompt controls.}
Of the 595 pages, 497 have both revisions timestamped by midnight on
$t_1-1$. On these pages we compare
$(0,-1)$ with $(-1,0)$ and rerun $(0,+1)$ versus $(+1,0)$.
Question-matching collection is later in the first pair but earlier in
the second, separating matching from earliest collection.
On the separate 128 pages, we repeat the question-day/next-day swap
with seven prompt variants. Three explain the timestamps in equivalent
wording: the original ``retrieved\_at is when this snapshot was fetched,''
and two rewrites describing the fetched snapshot's date or acquisition
timestamp. Four add an ignore-collection instruction, specify same-page
archive provenance, rename the field, or move it below the excerpt.
Appendix~\ref{app:n12-followup} gives the complete wording.
All retain both orders, fixed evidence and original-prompt controls
executed with the variants.

\paragraph{Content and spelling controls.}
A separate 128-record synthetic test uses invented entities and known
state changes: four relations (headquarters, coach, maintainer, operator)
times two formats (prose and key--value) times 16 records.
For example, a fictional company's headquarters changes from City A
in the older revision to City B in the newer one. The latter can state
``Its headquarters is in City B'' or \texttt{headquarters: City B}.
On the 128 Wikipedia pages, a second test changes only the question date from
ISO to English (e.g., January 16, 2025); metadata remains ISO.
Both use all five LLMs and both assignments/orders, with controls run
alongside the test conditions (Appendix~\ref{app:m12}).

\paragraph{Shared dates and comparison conditions.}
Shared-question-date assigns both revisions the same collection date, $t_1$.
Unknown keeps the fields and full definition but uses \texttt{unknown};
definition-only removes the fields while retaining their explanation;
omit also shortens that explanation.
The question-matched positive control instead assigns $r_0=t_0,r_1=t_1$.
Nondate controls replace the date field with \texttt{retrieval\_channel},
describing how rather than when a revision was obtained.
Its value, \texttt{cache} or \texttt{unknown}, is assigned, not observed.
The 128-page replication additionally crosses
the logged acquisition day $c$ and $c+7$, and tests shared $c+30$.
Only $(c,c)$ uses observed acquisition dates; shifted dates are assigned.

\paragraph{Reasoning and answer selection.}
Qwen3.5-35B answers the 128-page matching swap with two seeds in thinking
and sampled nonthinking modes, sharing sampling and output budgets.
An initial 32-page comparison is nested within this 128-page expansion;
the separate 599-page shared-date results appear in
Tables~\ref{tab:cohort599} and~\ref{tab:thinking-shared-interaction}.
Matching swaps also use two seeds with native reasoning in
Ministral-3-14B and Nemotron-3-Nano-30B-A3B
\citep{mistral2026ministral,nvidia2025nemotron}.
GPT-6 Astra and Claude Fable 5.1 \citep{openai2026astra,anthropic2026fable}
additionally answer the 128-page matching swap and closed-book questions
at high and low effort through independent API requests
(Appendix~\ref{app:n3-followup}).
They also answer both synthetic swap assignments at both efforts
(Appendix~\ref{app:synthetic-api}).
A separate A/B answer selection task asks the LLM to choose between the reference answers
for $t_0$ and $t_1$, balancing option labels and prohibiting abstention.
It tests shared-date conditions, not the matching swap.
Programmatic selection supplies the latest revision timestamped by the
question date. Exact inputs appear in
Appendix~\ref{app:prompts}.

\paragraph{Inference.}
Tables report paired effects and nominal 95\% predicate-cluster bootstrap
intervals after averaging source orders within page. An interval excluding
zero is not itself a multiplicity-adjusted decision.
Table~\ref{tab:adjusted-summary} reports adjusted decisions alongside the
complete-family definitions in Appendix~\ref{app:statistics}.
Secondary reanalyses are descriptive.

\section{Date Assignment Redirects Answers}
\label{sec:results}\label{sec:selection}\label{sec:date-results}

\paragraph{Date assignment changes answers.}
On 595 Wikipedia pages, assigning the question date $t_1$ as the collection
date of the older rather than the newer revision lowers later-reference
EM by 37.6--51.4 points across all five LLMs
(Table~\ref{tab:assignment-primary}).
For Qwen2.5-72B, EM changes from 77.1\% to 28.3\%; for Phi-4,
from 51.0\% to 3.7\%. Earlier-reference EM ($t_0$) rises 37.6--49.8
points with revision texts and timestamps unchanged; only collection-date
assignment changes.
Earlier-answer-only inclusion also rises.
\begin{table*}[t]
\centering\footnotesize
\setlength{\tabcolsep}{3pt}
\begin{tabular}{lrrr}
\toprule
LLM & \shortstack{Earlier-reference\\EM ($t_0$)} & \shortstack{Earlier answer only\\($t_0$)} & \shortstack{Later-reference\\EM ($t_1$)} \\
\midrule
\multicolumn{4}{l}{\textit{595 pages, 113 predicates, raw display}} \\
Qwen2.5-7B & +38.7 [+29.9,+47.0] & +41.9 [+34.8,+48.9] & -40.1 [-49.4,-30.6] \\
Phi-4 & +45.1 [+31.6,+58.3] & +65.2 [+56.8,+73.5] & -47.3 [-62.2,-32.0] \\
Mistral-24B & +49.8 [+39.0,+60.2] & +51.3 [+40.9,+61.2] & -51.4 [-62.6,-39.9] \\
Qwen3.5-35B & +37.6 [+30.6,+44.4] & +39.5 [+33.5,+45.2] & -37.6 [-44.1,-30.9] \\
Qwen2.5-72B & +44.9 [+34.0,+57.4] & +48.3 [+39.2,+59.4] & -48.7 [-60.2,-36.9] \\
\midrule
\multicolumn{4}{l}{\textit{128 pages, 45 predicates, plain display}} \\
Qwen2.5-7B & +36.7 [+27.2,+45.7] & +40.6 [+32.4,+48.4] & -38.3 [-48.0,-28.0] \\
Phi-4 & +40.2 [+26.8,+53.1] & +68.8 [+58.1,+78.0] & -43.0 [-56.9,-28.9] \\
Mistral-24B & +46.1 [+33.1,+60.6] & +53.5 [+41.9,+66.0] & -47.7 [-62.7,-33.7] \\
Qwen3.5-35B & +41.0 [+29.8,+52.7] & +45.7 [+37.0,+54.4] & -40.6 [-51.7,-29.6] \\
Qwen2.5-72B & +47.7 [+33.7,+63.7] & +56.2 [+45.2,+69.7] & -53.1 [-67.8,-38.7] \\
\bottomrule
\end{tabular}
\caption{Question-date matching swap, $(o{=}0,n{=}1)-(o{=}1,n{=}0)$, for $t_1$ questions. Values are percentage-point changes against $t_0$/$t_1$ references with nominal 95\% predicate-cluster intervals. The 128 pages are from the separate 596-page cohort, disjoint from the 595 pages. Scores average both revision orders per page. Earlier-answer-only inclusion uses the same text normalization as EM in both cohorts; follow-ups additionally use the Unicode word-matching rule described in Appendix~\ref{app:data}. All EM levels and proximity intervals: Appendix~\ref{app:assignment-replication}.}
\label{tab:old-answer-swaps}
\label{tab:assignment-primary}
\end{table*}

\paragraph{Near-question proximity swaps.}
Assigning the older revision the collection date closer to $t_1$
lowers EM by 28.2--38.9 points
in Phi-4, Mistral-24B, Qwen3.5-35B and Qwen2.5-72B;
closeness and earlier collection coincide here.
Qwen2.5-7B instead changes by $+1.2$ points (95\% interval $[-0.5,2.8]$),
despite losing 40.1 points in the question-date matching swap.
Table~\ref{tab:proximity-intervals} gives both cohorts' intervals.
The much smaller distant swaps (Section~\ref{sec:generalization}) argue
against a distance-independent earliest-collection rule.

\paragraph{Replication on 128 separate pages.}
On 128 Wikipedia pages that do not overlap with the 595 pages used above,
question-date matching swaps lower EM by 38.3--53.1 points in all five LLMs.
Proximity swaps again affect the other four LLMs
($-28.5$ to $-37.1$ points), but not Qwen2.5-7B
($-0.4$).
Table~\ref{tab:essential-grid-em} reports all 18 same-run levels,
including unknown, definition-only and omission rerun with the swaps.
These disjoint pages were previously used for other conditions.

\paragraph{Sensitivity is not a symmetric nearest-date rule.}
A newer-revision collection date closer to $t_1$ does not uniformly
improve on shared dates:
Qwen2.5-7B scores
50.0\% versus 55.9\%, and Qwen2.5-72B 61.1\% versus 68.9\%
(Table~\ref{tab:cue-groups}). Likewise, an informative question-matched
cue improves EM for questions about $t_0$ over omission by 13.5--24.8 points on
596 pages (Table~\ref{tab:earlier-qmatch}):
revision timestamps alone do not ensure correct selection.
Cue competition here describes behavior, not internal weights.

\paragraph{Dates on or before the question retain the effect.}
On 497 pages, $(0,-1)$ versus $(-1,0)$ makes the revision collected on question date
$t_1$ the later, not earlier, collection. EM still falls 29.0--39.6 points
across all five LLMs when the older revision receives $t_1$
(Table~\ref{tab:n1-main}).
EM against the earlier answer rises 22.5--37.1 points. A preference for the earliest
collection therefore cannot alone explain the effect. Direction still
matters: losses attenuate by 17.9, 12.4 and 17.4 points in Phi-4,
Mistral-24B and Qwen2.5-72B relative to after-question controls
run alongside these conditions.
\begin{table*}[t]
\centering\footnotesize\setlength{\tabcolsep}{3pt}
\begin{tabular}{lllll}
\toprule
LLM & After: later EM & Before: later EM & Before: earlier EM & Before $-$ after \\
\midrule
Qwen2.5-7B & -40.4 [-50.6,-30.1] & -39.0 [-48.2,-29.5] & +37.1 [+27.9,+46.3] & +1.4 [-1.3,+4.5] \\
Phi-4 & -46.9 [-61.5,-31.6] & -29.0 [-38.4,-19.1] & +22.5 [+14.2,+30.9] & +17.9 [+11.7,+24.2] \\
Mistral-24B & -52.0 [-63.8,-40.3] & -39.6 [-51.6,-27.7] & +32.4 [+22.5,+42.6] & +12.4 [+8.7,+15.6] \\
Qwen3.5-35B & -38.5 [-45.7,-31.4] & -34.3 [-42.6,-26.0] & +33.6 [+24.8,+42.5] & +4.2 [+0.7,+7.2] \\
Qwen2.5-72B & -49.6 [-61.9,-36.9] & -32.2 [-40.8,-23.6] & +29.6 [+21.4,+39.0] & +17.4 [+12.4,+23.3] \\
\bottomrule\end{tabular}
\caption{Changes within one execution for questions about $t_1$ on 497 eligible Wikipedia pages (103 predicate groups). After is $(0,+1)-(+1,0)$; before is $(0,-1)-(-1,0)$, always older-matching minus newer-matching assignment. Later/earlier EM uses the $t_1$/$t_0$ reference. The last column compares later-reference EM effects. Units are percentage points; brackets give nominal 95\% predicate-cluster intervals. Full levels: Appendix~\ref{app:n12-followup}.}
\label{tab:n1-main}
\end{table*}

\paragraph{Instruction and provenance controls.}
On 128 pages, all seven prompt variants retain large question-date
matching swap losses in every LLM (Table~\ref{tab:n2-main}). Telling LLMs not to select
by collection time leaves 37.9--50.8-point losses; explicitly stating
that the snippets are archived revisions leaves 42.2--57.8-point losses.
Earlier-answer content rises in both conditions. Neither an unstated
live-acquisition interpretation nor one field name, position or definition alone explains the
effect. Archive provenance increases Mistral's penalty by 10.2 points.
Appendix~\ref{app:n12-followup} reports all variant-minus-original interactions.
\begin{table*}[t]
\centering\footnotesize\setlength{\tabcolsep}{3pt}
\begin{tabular}{llllll}
\toprule
Prompt contrast & Qwen2.5-7B & Phi-4 & Mistral & Qwen3.5-35B & Qwen2.5-72B \\
\midrule
Original & \shortstack{-37.9 / +36.3\\{[-48.1,-28.0]}} & \shortstack{-42.6 / +40.2\\{[-57.3,-27.9]}} & \shortstack{-47.7 / +45.7\\{[-63.1,-33.5]}} & \shortstack{-39.5 / +39.5\\{[-50.6,-28.4]}} & \shortstack{-52.7 / +47.3\\{[-67.0,-38.6]}} \\
Definition 1 & \shortstack{-41.0 / +38.3\\{[-51.3,-30.1]}} & \shortstack{-46.9 / +45.3\\{[-61.4,-32.0]}} & \shortstack{-54.7 / +48.0\\{[-68.0,-41.5]}} & \shortstack{-44.9 / +44.1\\{[-57.2,-32.5]}} & \shortstack{-57.4 / +52.0\\{[-71.3,-42.6]}} \\
Definition 2 & \shortstack{-41.8 / +38.3\\{[-51.9,-30.5]}} & \shortstack{-46.5 / +44.1\\{[-60.4,-31.8]}} & \shortstack{-44.5 / +43.4\\{[-59.0,-31.4]}} & \shortstack{-42.2 / +42.2\\{[-53.8,-30.5]}} & \shortstack{-54.7 / +50.4\\{[-70.6,-39.7]}} \\
Ignore collection & \shortstack{-42.2 / +39.5\\{[-52.2,-30.9]}} & \shortstack{-43.0 / +41.0\\{[-57.2,-28.1]}} & \shortstack{-47.3 / +43.0\\{[-60.5,-34.3]}} & \shortstack{-37.9 / +38.3\\{[-48.2,-28.4]}} & \shortstack{-50.8 / +46.1\\{[-65.0,-37.7]}} \\
Archive provenance & \shortstack{-43.4 / +41.0\\{[-54.0,-32.2]}} & \shortstack{-45.3 / +45.7\\{[-59.8,-29.9]}} & \shortstack{-57.8 / +55.9\\{[-71.3,-44.4]}} & \shortstack{-42.2 / +41.8\\{[-54.0,-31.0]}} & \shortstack{-52.3 / +47.3\\{[-67.5,-38.5]}} \\
Field name & \shortstack{-43.8 / +42.2\\{[-54.5,-32.1]}} & \shortstack{-44.9 / +42.6\\{[-59.4,-29.9]}} & \shortstack{-48.4 / +44.1\\{[-62.3,-34.9]}} & \shortstack{-46.5 / +46.5\\{[-59.8,-34.1]}} & \shortstack{-57.0 / +50.8\\{[-71.2,-42.4]}} \\
Field position & \shortstack{-50.4 / +46.5\\{[-62.0,-38.3]}} & \shortstack{-35.9 / +31.2\\{[-50.0,-21.2]}} & \shortstack{-54.7 / +51.2\\{[-68.6,-40.3]}} & \shortstack{-45.7 / +44.5\\{[-58.8,-33.2]}} & \shortstack{-49.2 / +41.8\\{[-62.9,-35.9]}} \\
\bottomrule\end{tabular}
\caption{Question-date matching swaps for $t_1$ questions on the same 128 Wikipedia pages (45 predicate groups). Cells show later-reference ($t_1$) / earlier-reference ($t_0$) EM changes, older-matching minus newer-matching, in percentage points. Brackets give nominal 95\% predicate-cluster intervals for later-reference EM; earlier-reference intervals: Table~\ref{tab:n2-old-em}. Levels and prompt interactions: Appendix~\ref{app:n12-followup}.}
\label{tab:n2-main}
\end{table*}

\paragraph{Redirection with fictional answers.}
On 128 invented records, question-date matching swaps lower later-reference
EM ($t_1$) by 92.2--95.7 points across all five LLMs
(Table~\ref{tab:m12-main}); earlier-reference EM ($t_0$) rises 91.4--96.1
points. Without collection fields,
two-version EM is 96.9--100.0\%, and newer-only EM is 93.0--100.0\%
(Table~\ref{tab:m1-levels}). Both prose and key--value styles retain
large losses. Near-ceiling controls allow larger absolute losses
than Wikipedia controls. These minimal records also lack internal
history cues such as the former-team field in Appendix~\ref{app:prompts};
whether such cues explain part of the gap is untested. Redirection needs no remembered
real-world answers.

\paragraph{Different question-date spellings.}
Changing only the question date to English month-name spelling leaves
31.6--53.1-point swap losses in all five LLMs, with increased earlier-reference EM
(Tables~\ref{tab:m12-main} and~\ref{tab:m2-levels}). All five losses remain
significant across the complete 60-test family
(Table~\ref{tab:adjusted-summary}).
Phi-4's loss decreases by 11.3 points; the other four changes remain
unresolved after correction.
Thus identical strings are unnecessary for the effect.
\begin{table}[t]
\centering\footnotesize\setlength{\tabcolsep}{2pt}
\begin{tabular}{llll}
\toprule
LLM & Synthetic & English date & \shortstack{English\\minus ISO} \\
\midrule
Qwen2.5-7B & \shortstack{-95.7\\{[-98.0,-93.4]}} & \shortstack{-34.8\\{[-45.2,-24.2]}} & \shortstack{+2.7\\{[-2.9,+8.3]}} \\
Phi-4 & \shortstack{-94.5\\{[-97.7,-91.0]}} & \shortstack{-31.6\\{[-42.1,-20.6]}} & \shortstack{+11.3\\{[+4.3,+19.1]}} \\
Mistral & \shortstack{-94.1\\{[-96.9,-91.4]}} & \shortstack{-53.1\\{[-65.7,-38.9]}} & \shortstack{-5.5\\{[-13.4,+1.6]}} \\
Qwen3.5-35B & \shortstack{-92.6\\{[-95.3,-89.5]}} & \shortstack{-35.9\\{[-46.0,-25.9]}} & \shortstack{+3.5\\{[-0.9,+7.5]}} \\
Qwen2.5-72B & \shortstack{-92.2\\{[-95.3,-89.1]}} & \shortstack{-46.5\\{[-59.1,-34.7]}} & \shortstack{+6.6\\{[+2.7,+11.0]}} \\
\bottomrule\end{tabular}
\caption{Changes in later-reference ($t_1$) EM after swapping collection dates, in percentage points. Synthetic uses 128 fictional records; English date uses 128 Wikipedia pages with month-name question dates. The last column is the English-date minus same-run ISO-date swap effect. Brackets: nominal 95\% record intervals for synthetic data and predicate-cluster intervals for Wikipedia pages. Baselines: Appendix~\ref{app:m12}.}
\label{tab:m12-main}
\end{table}

\section{Date Roles and Answer Selection}
\label{sec:control-results}\label{sec:validity}\label{sec:cache}
\label{sec:answer-measurement}

\paragraph{Displaying fields with observed dates.}
On 128 pages, displaying $(c,c)$ yields 3.5--15.2-point lower later-reference
EM than omission (Table~\ref{tab:observed-dates}). This jointly changes
fields, their definition and date values. The single acquisition day $c$
is 178--755 days after $t_1$. After expanded correction, omission contrasts
remain in Phi-4, Mistral-24B and Qwen2.5-72B; definition-only contrasts
remain in Phi-4 and Mistral-24B. Unknown-minus-omission remains in Phi-4
and Mistral-24B; the further $(c,c)$-minus-unknown change, $-2.3$ to $-6.6$
points, is unresolved in all five, so a separate date-value effect is not established.

\paragraph{Sensitivity extends beyond retrieval semantics.}
Publication, last-modified and explicitly irrelevant administrative dates
also change answers in the 595-page plain-display study
(Table~\ref{tab:roles-compact}), using shared-minus-question-matched
comparisons rather than the main swap.

\paragraph{Answer content and comparator choice.}
On 411 pages whose input contains both reference answers, later-reference
EM falls 44.6--59.4 points while earlier-reference EM rises 44.4--57.5
(Table~\ref{tab:alias-subset}).
Shared-date changes instead mix content and form. Missing-value choice
changes the baseline: Phi-4's later EM is 4.4 points lower with unknown
than with definition-only, explaining its $-12.9$ versus $-17.3$ shared-date
contrasts (Table~\ref{tab:review-baselines}). Separately, a nondate placebo
gives $-1.5$ to $+0.6$ points; direct shared-minus-channel-cache effects
remain negative in four LLMs, with the 72B interval including zero.
In A/B answer selection, Phi-4 alone has corrected evidence of an
earlier-answer shift at both question times. Mistral's later interval
includes zero, and Qwen2.5-7B's earlier direction reverses;
Table~\ref{tab:ab-clusters} reports all four LLMs.

\section{Reasoning and Scope}
\label{sec:generalization}

\paragraph{Matched Qwen3.5 thinking.}
On 128 pages with two seeds, thinking reduces the EM swap penalty from
$-30.1$ to $-11.1$ points (Table~\ref{tab:remaining-reasoning}).
The $+18.9$-point interaction has a predicate-cluster interval
$[11.1,26.5]$.
Both seeds attenuate, with residual thinking losses of
8.2 and 14.1 points (Table~\ref{tab:remaining-seeds}). Earlier-reference EM increases by 30.5 points without
thinking and 11.1 with it. Both modes share sampling settings and output
caps; greedy outputs are not the control. The same 128-page greedy penalty
is 40.6 points; this difference cannot be attributed to thinking alone.
\begin{table*}[t]
\centering\scriptsize
\setlength{\tabcolsep}{3pt}
\begin{tabular}{llrrrr}
\toprule
LLM & Mode & \shortstack{Newer revision\\matches $t_1$} & \shortstack{Older revision\\matches $t_1$} & \shortstack{$\Delta$ later-reference\\EM ($t_1$)} & \shortstack{$\Delta$ earlier-reference\\EM ($t_0$)} \\
\midrule
Qwen3.5-35B & sampled & 59.6 & 29.5 & -30.1 [-37.1,-23.1] & +30.5 [+23.5,+37.8] \\
Qwen3.5-35B & thinking & 72.1 & 60.9 & -11.1 [-16.3,-7.0] & +11.1 [+7.0,+16.3] \\
Ministral-3-14B & thinking & 56.1 & 43.0 & -13.1 [-18.1,-8.1] & +13.3 [+8.5,+18.1] \\
Nemotron-3-Nano & thinking & 65.2 & 13.5 & -51.8 [-64.3,-37.8] & +48.8 [+35.4,+61.6] \\
\bottomrule
\end{tabular}
\caption{Question-date matching swaps for $t_1$ on the same 128 Wikipedia pages (45 predicate groups). Matching columns give later-reference EM (\%); changes are older-matching minus newer-matching, in percentage points with nominal 95\% predicate-cluster intervals. Scores average two seeds and both revision orders. Qwen modes share sampling; cross-family comparisons do not isolate reasoning. Settings and tests: Appendix~\ref{app:remaining-local}.}
\label{tab:remaining-reasoning}
\end{table*}

\paragraph{Native reasoners vary.}
On the 128-page, two-seed question-date matching test, Ministral-3-14B and
Nemotron-3-Nano lose 13.1 and 51.8 EM points, respectively.
Earlier-reference EM rises by 13.3 and 48.8 points, with the same direction
in both seeds (Table~\ref{tab:remaining-reasoning}). Sensitivity therefore
extends beyond the five nonthinking LLMs but varies by configuration.

\paragraph{API tests on Wikipedia revisions.}
On the same 128 pages, 2,560 independent API responses cross high/low
effort with swap and closed-book controls (Table~\ref{tab:n3-main}), without
author system messages, added current dates, history or tools.
GPT-6 Astra/Claude Fable 5.1 have 21.9/7.8\% high-effort closed-book EM.
On their 100/118 EM-miss pages, high/low swap effects are
$-2.0/-1.0$ and $-0.4/-2.5$ points; intervals establish neither loss nor
effort interaction. The subsets condition on observed closed-book responses
(Appendix~\ref{app:n3-followup}).
\begin{table*}[t]
\centering\footnotesize\setlength{\tabcolsep}{3pt}
\begin{tabular}{llllll}
\toprule
API LLM & Effort & CB EM & Swap: all 128 & Miss $n$ & Swap: CB-high misses \\
\midrule
GPT-6 Astra & high & 21.9 & -1.2 [-4.8,+1.5] & 100 & -2.0 [-6.4,+0.6] \\
GPT-6 Astra & low & 18.0 & -0.8 [-2.2,+0.0] & 100 & -1.0 [-2.7,+0.0] \\
Claude Fable 5.1 & high & 7.8 & -0.4 [-1.9,+0.9] & 118 & -0.4 [-2.0,+1.0] \\
Claude Fable 5.1 & low & 7.0 & -2.3 [-6.3,+1.1] & 118 & -2.5 [-6.4,+0.9] \\
\bottomrule
\end{tabular}
\caption{Contemporaneous high/low API comparison. CB is closed-book reference EM (\%). Swap: older-matching minus newer-matching EM percentage points (nominal 95\% predicate-cluster intervals). Each LLM\textquotesingle s high-effort closed-book EM misses define a subset reused at both efforts, not proof of absent knowledge. Levels and conditional sensitivity: Appendix~\ref{app:n3-followup}.}
\label{tab:n3-main}
\end{table*}

\paragraph{Synthetic API answer content.}
Both API LLMs achieve 100\% high-effort EM under both assignments
on 128 invented records (Table~\ref{tab:synthetic-api}). At low effort,
GPT-6 Astra loses 0.4 points and Claude Fable 5.1 loses 14.1.
Claude's loss and its increase over high effort survive correction
(Table~\ref{tab:adjusted-summary}).
Of Claude's 256 responses under the lower-EM assignment, 220 exactly match
the later answer, 14 include it with additional text but no earlier answer,
21 mention both answers, and one contains only the earlier answer.
Overall, 99.6\% include the later answer, versus 0.4\% earlier-answer-only.
These longer or hedged answers differ from the earlier-answer shift
in local LLMs; inclusion alone does not establish unambiguous selection.

\paragraph{Distant-date proximity.}
With dates months after both questions, $(c,c+7)$ versus $(c+7,c)$ swaps
give 0.0--6.2 EM points for $t_0$ questions and 0.0--2.7 for $t_1$ questions
(Appendix~\ref{app:remaining-local}). On 262 Temporal Wiki pages
\citep{ozer2025temporal}, analogous
proximity effects are $-0.2$ to $+0.8$ points across three LLMs and
both question years. These swaps do not reproduce near-question losses,
arguing against a date-distance-independent preference for the earliest
collection. They differ from displaying a shared observed date versus
removing its fields (Table~\ref{tab:observed-dates}).
Temporal Wiki's year-level questions and later revisions prevent a
chronology-valid exact question-date match; that main contrast remains
untested in a second natural dataset.

\paragraph{Revision selection.}
On 595 pages, selecting the latest eligible revision before
generation improves EM over two-version omission by 15.0--30.9 points
for $t_0$ questions (Table~\ref{tab:earlier-selection}). Revision competition can be substantial
even without collection fields; date assignment is an additional influence.
The rule assumes known source identity and can discard later corrections.

\paragraph{A practical metadata audit.}
Keep revision texts and timestamps fixed, swap chronology-valid collection dates
in both orders, and score agreement and earlier-answer content. Test both
sides of the question and archive provenance: the relevant revision is
present, yet metadata can change answers.

\section{Related Work}
\label{sec:related}

\paragraph{Evidence conflicts and temporal validity.}
Conflicting-source studies \citep{xu2024conflicts,chen2022conflicts},
evidence receptiveness \citep{wan2024convincing,xie2024chameleon}, and
changing facts \citep{longpre2021conflicts,marjanovic2024dynamicqa,wallat2026facts}
motivate version-sensitive reading. ClashEval studies retrieved content
versus model priors \citep{wu2024clasheval}; CONFLICTS labels retrieval
conflicts \citep{cattan2025conflicts}.
ConflictBank includes temporal discrepancies among several knowledge-conflict
types \citep{su2024conflictbank}. Multiple in-context updates can bias retrieval
toward earlier rather than current values \citep{qiao2026updates}.
We instead fix both revision texts and timestamps, changing a separate
collection-date field.

\citet{ahmed2026stale} vary evaluation time against fixed historical evidence
and explicit validity-end boundaries. We instead fix question time and both
revisions while changing collection metadata. Our archive condition clarifies
revision provenance, not a verified end to factual validity.

\paragraph{Preselection and source credibility.}
HoH, Re3 and FRESCO study outdated retrieval and reranking
\citep{ouyang2025hoh,cao2026re3,an2026fresco}; VersionRAG represents version
lineage \citep{huwiler2025versionrag}, and ContextNest evaluates governed
selection under stale-version exposure \citep{sulpovar2026contextnest}.
Publication-date injection affects reranking \citep{fang2025recency}; our
field-role result shows that stated date roles need not determine their
effects on answers. We isolate date assignment between two fixed revisions
at the answering stage; date injection itself is not new.
TempRALM and MRAG use temporal selection/ordering
\citep{gade2024tempralm,zhang2025mrag}. Credibility-aware generation learns
source weighting \citep{pan2024credibility}. We test the answering LLM
with fixed evidence.

\paragraph{Surface cues and controlled diagnostics.}
Controlled heuristic and distraction tests
\citep{jia2017adversarial,mccoy2019hans,shi2023distracted} motivate separating
task-relevant evidence from surface cues.
DateLogicQA examines date-format bias \citep{bhatia2024datelogic}; our prompt
controls keep conflicting revision evidence fixed.
TempReason and Test of Time assess temporal reasoning
\citep{tan2023tempreason,fatemi2025testoftime}, whereas system-date effects
\citep{sanz2026dating} change global model context.
Option-ID/order bias \citep{zheng2024selectors,pezeshkpour2024order} and
context-position sensitivity \citep{liu2024lost} motivate our balanced
candidate labels and evidence orders.
Prompt-format sensitivity and multi-prompt evaluation
\citep{sclar2024formatting,mizrahi2024multiprompt} motivate our changes to
timestamp definitions and instructions. \citet{cao2022timeprompt} inject
textual or continuous timestamp prompts for generation; we distinguish
collection dates from revision timestamps in otherwise fixed conflicting context.

\paragraph{Temporal benchmarks and scoring.}
Time-aware knowledge models \citep{dhingra2022time,zhao2024clock} and dated
QA benchmarks \citep{chen2021timeqa,zhang2021situated,liska2022streaming,kasai2023realtime,vu2024freshllms,lin2025dynaquest}
study changing answers. Dynamic-TempLAMA \citep{margatina2023dynamic} and
TemporalWiki \citep{jang2022temporalwiki} evaluate temporal knowledge updates.
Our historical article-text diagnostic uses the distinct dataset of
\citet{ozer2025temporal}, not Jang et al.'s TemporalWiki.
TIDE tests expert-verified amendment validity \citep{sobhani2026tide}.
Selective QA and sufficiency \citep{kamath2020selective,wen2025limits,joren2025sufficient}
motivate separating reference agreement from factual error.

\section{Conclusion}

\textbf{RQ1:} Collection-date swaps with fixed revision texts and timestamps
redirect answers toward earlier answers on nonoverlapping Wikipedia page
sets and fictional records.
\textbf{RQ2:} Losses persist before and after question date $t_1$, across
seven prompt variants and English question dates. Earlier-answer content
increases. Neither earliest
collection nor identical date strings is required.
\textbf{RQ3:} Matched thinking reduces Qwen3.5's penalty to 11.1
points; native reasoners differ. API collection-date swaps on Wikipedia
revisions produce small EM changes. Claude's synthetic low-effort loss
mainly reflects longer answers, not earlier-answer selection.
Preselection addresses separate revision competition.
Large swap-induced losses concern matching and nearby collection dates;
swaps between distant dates yield much smaller changes. Displaying a
shared observed acquisition date lowers agreement relative to omission
in three of five LLMs after correction, a joint field-and-value comparison
rather than an isolated field-presence effect.
Do not assume metadata is neutral: assess evidence,
date assignment and answer form separately.

\FloatBarrier
\clearpage
\setlength{\dbltextfloatsep}{20pt plus 2pt minus 4pt}
\setlength{\textfloatsep}{20pt plus 2pt minus 4pt}
\section*{Limitations}

\paragraph{Revision recovery and lexical outcomes.}
EM measures agreement with dated benchmark references, not independently
adjudicated historical truth. An edit may describe an event or correct an
earlier statement; their proportions are not established. Alias inclusion
checks lexical availability, not entailment. The both-alias sensitivity
retains the primary sample. A/B answer selection supplies
reference answers and forbids abstention, and has no matching-swap condition.

\paragraph{Configurations and transfer.}
The main experiment uses two English Wikipedia revisions per page,
with at most 4,003 characters each. Assigned-date losses and the mirror/cache
scenarios do not estimate live-system error rates. Only one acquisition
day is observed.
Distant-date and external proximity swaps do not reproduce the large
near-question swap losses; this does not make displaying distant metadata neutral.
A past-side proximity swap, $(-1,-7)$ versus $(-7,-1)$, was not run.
After-question proximity alone conflates closeness with earlier collection.
The question-date matching swap remains untested in a second natural
QA dataset. The fictional-record replication has known generated truth
but only eight fixed templates; its intervals condition on those templates.
Under $v_0\leq t_0<v_1\leq t_1$, assigning $r_1=t_0$ is invalid;
the reported earlier-question proximity control is not a matching-swap
counterpart.

\paragraph{Prompts, rendering and mechanisms.}
Seven prompts do not represent a prompt population; rendering and cohort
both differ across some studies. Field-role results use shared versus question-matched
dates, not the matching swap. The latter has not been crossed with removal
of revision timestamps, so cue competition is a behavioral description,
not an identified internal weighting rule. Cross-format persistence does
not establish a unique internal calendar
algorithm or exclude all lexical features. Explicit instructions and
archive provenance do not eliminate the tested swaps; this does not rule
out other prompting remedies. Unknown is one missing-value token and
cache one concrete placebo, with imperfect semantic/length matching.

\paragraph{Exposure and execution.}
The three principal cohorts are page-disjoint, but follow-ups reuse pages
and were informed by earlier results. Local plans precede their own
outputs, not all preceding research, and are not public preregistration.
Recorded exclusions cannot rule out pretraining exposure. External
diagnostics retain disclosed development overlap. The 32-page reasoning
study is nested in the 128-page expansion, not a second replication.
Matched thinking concerns one Qwen family and two seeds; decoding
dependence is reported in the main text. Cross-family differences do not
isolate reasoning. Incomplete final answers remain in the denominator,
with completion bounds distinct from confidence intervals.

\paragraph{API LLMs and practical mitigation.}
Public model identifiers are fixed, but immutable backends and provider
instructions are unobservable. Each order/effort has one execution.
Closed-book EM misses are not proof of absent knowledge; conditional
intervals fix the observed subset. Small effects at two effort settings
do not establish equivalence or robustness in other settings. GPT-6 Astra has no
reasoning-off setting and Claude Fable 5.1 uses always-on adaptive thinking
\citep{openai2026astra,anthropic2026fable}. These API comparisons therefore
cannot separate model identity from the presence of reasoning. Programmatic revision
selection assumes known lineage and question time and can discard later
corrections. It is a conditional revision-selection policy, not a general
solution to temporal factuality or a measured latency/energy reduction.

\section*{Ethical Considerations}

No new human participants are involved. Wikipedia can contain personal
names or inaccurate statements; outputs are not verified facts. Assigned
dates are labeled as interventions. English Wikipedia text and
DynaQuest are reused under CC BY-SA 4.0, retaining source attribution
and revision links. The Temporal Wiki repository of \citet{ozer2025temporal}
is MIT-licensed; its underlying Wikipedia text retains Wikipedia's terms.
Original reproduction code uses the MIT license;
third-party data and models retain their own terms. Supplements exclude
credentials, accounts and model weights. Execution settings and API call
counts appear in Appendix~\ref{app:runtime}.

\section*{Use of AI Assistants}

AI assisted coding, manuscript review, English editing, and writing.

\clearpage
\appendix
\setlength{\dbltextfloatsep}{10pt plus 2pt minus 2pt}
\setlength{\textfloatsep}{10pt plus 2pt minus 2pt}
\maxdeadcycles=1000
\setcounter{dbltopnumber}{6}
\renewcommand{\dbltopfraction}{0.9}
\renewcommand{\textfraction}{0.08}
\raggedbottom
\section{Samples, Rendering and Measurement}
\label{app:data}\label{app:statistics}

The separate \emph{Extended Technical Results} retains exhaustive
comparisons, alternative intervals and earlier diagnostics.

\paragraph{Cohort construction.}
Each DynaQuest cohort starts with a 600-page sample, without replacement
or answer-presence filtering. From 11,441 eligible pairs, one missing
revision, three nonforward pairs and one previously used identity leave
595 pages for the main experiment. A fixed hash of page identities selects the second sample
from 10,289 pairs; three chronology violations and one prior identity
leave 596. Excluding those cohorts and pages used in earlier experiments leaves 9,201 pairs
for the third sample, with one chronology violation leaving 599.
The three cohorts are disjoint by canonical page ID. Exclusions refer to
our recorded use, not pretraining. The 128-page date-swapping sample is a fixed
subset of 596; the separate 128-page A/B answer selection sample is from 595.
Plans precede each follow-up's responses, but the pages and earlier
conditions had already been studied.

\begin{table*}[t]
\centering\footnotesize\setlength{\tabcolsep}{4pt}
\begin{tabularx}{\textwidth}{lrllX}
\toprule
Study & Pages & Rendering & LLMs & Relationship \\
\midrule
Main date assignments & 595 & Raw & Core five & Original cohort \\
Question-day/previous-day swaps & 497 & Raw & Core five & All eligible pages of 595 \\
Roles / display controls & 595 & Plain / both & Core five & Same original pages \\
Baseline decomposition & 596 & Plain & Core five & Disjoint second cohort \\
Grid / seven prompts & 128 & Plain & Core five & Fixed subset of 596 \\
Expanded thinking & 128 & Plain & Qwen3.5-35B & Same subset; two seeds \\
Native reasoning / API & 128 & Plain & Four added LLMs & Same subset; not new cohorts \\
Shared-date replication & 599 & Plain & Core five & Disjoint third cohort \\
A/B answer selection & 128 & Raw & Four, not 72B & Different subset of 595 \\
External proximity & 262 & Native article text & Three & Temporal Wiki \citep{ozer2025temporal}, nine relations \\
\bottomrule
\end{tabularx}
\caption{Sample and rendering map. Raw preserves wiki markup; plain preserves its words/numbers while changing delimiters and whitespace. All contrasts hold evidence rendering fixed. Nested samples are not independent cohorts. The initial 32-page reasoning-setting comparison is included in the expanded 128-page experiment.}
\label{tab:cohort-map}
\end{table*}

\paragraph{Evidence and date provenance.}
BM25 uses only the question to select up to four 1,000-character blocks
per revision (at most 4,003 joined characters) as LLM evidence. Raw display
preserves the retrieved wiki text; plain display
removes double template/link delimiters, maps pipes to newlines and equals
to colons, and normalizes whitespace while preserving every Unicode
word/number token in order. It does not execute MediaWiki templates.
The rendering column in Table~\ref{tab:cohort-map} is verified from archived
task records. In the 595-page study that runs both displays together, raw/plain
rendering changes Mistral's later omit/question-matched EM from
73.9/78.6\% to 60.0/66.3\%, and Qwen2.5-72B's from
76.6/77.3\% to 68.2/69.7\%. These are paired display-study levels,
not substituted baseline answers for the later grid execution. Consequently,
between-cohort absolute levels are not estimates of cohort difficulty.

The two distant-date executions use identical messages for all 768
overlapping later-question inputs. Greedy reruns differ by at most
1.6 EM points in levels and 2.3 points in swap contrasts
(unrounded maximum 2.34375). Greedy decoding therefore does not ensure
bitwise reproducibility across separately batched executions; contrasts
always use their own execution's controls. The main grid and display
study likewise have identical omission prompts: later EM differs by
0.08 points in Mistral and 0.25 in Qwen2.5-72B across these executions.

The 596-page cohort has 39 question dates, September 2024--April 2026,
on the first or sixteenth of a month. These are benchmark dates, not
independently verified event times. Its 1,192 versions were acquired on
October 11, 2026; native hashes and BM25 spans replay offline. The batch
is 193--770 days after earlier questions and 178--755 after later ones.
Subsequent experiments reuse those sources; they do not constitute new
acquisitions. Only the logged day is observed; all shifted dates are assigned.

\paragraph{Scores.}
EM uses NFD Unicode, lowercase, ASCII-punctuation and article removal,
then whitespace normalization, accepting any nonempty reference alias.
Questions about $t_0$/$t_1$ use earlier/later references. F1 is maximum token F1
over aliases. Literal abstention is an exact whole-response match to
I don't know, I do not know, unknown, not enough information, insufficient
information, not specified, or not provided; it is not a semantic refusal
classifier. Earlier/later-only inclusion requires a whole-token alias from
one reference set but none from the other. The main cross-cohort content
comparison uses EM-style normalization in both cohorts; frozen follow-up
analyses use NFKD casefolded word sequences and retain their separate
scores. Neither aliases nor exact matches establish entailment.

\paragraph{Intervals and multiplicity.}
Page effects average orders and, where applicable, seeds. Predicate
bootstraps resample complete groups with paired pages: 10,000 draws,
follow-up seed 20261012. Tables~\ref{tab:inference-map}
and~\ref{tab:adjusted-summary} list original and post hoc correction
families and summarize adjusted results; effect intervals remain nominal.
The alias subset and rendering audit are post hoc descriptive analyses.
\begin{table*}[t]
\centering\footnotesize
\begin{tabularx}{\textwidth}{lX}
\toprule
Study & Complete correction scope retained in the technical record \\
\midrule
Original / role diagnostics & Original planned families and their 877-test union; retrospective content descriptions separate. \\
Main-grid swaps (post hoc) & 5 matching and 5 proximity tests; family 10 (Table~\ref{tab:proximity-intervals}). \\
Observed dates (post hoc) & Original 2 contrasts $\times$ 5 LLMs = 10; expanded 4 contrasts $\times$ 5 LLMs = 20 (Table~\ref{tab:observed-dates}). \\
128-page grid / nested modes & 51 EM tests plus 102 exclusive-alias tests; union 153. \\
Expanded reasoning / proximity & 6 native-reasoner and 66 mode/proximity tests; union 72. \\
Past-side / seven prompts / API & 80 local EM, 160 local alias, 22 API tests; union 262. \\
Shared-date controls & 20 effects plus 70 decompositions; union 90. Post hoc direct controls retain a deduplicated 120-test union. \\
A/B answer selection & 4 LLMs $\times$ 2 times $\times$ 5 contrasts = 40. \\
Synthetic / question spelling & 5 synthetic and 15 spelling EM tests; 40 exclusive-alias tests; union 60. \\
Synthetic API & 6 EM and 12 exclusive-alias tests; union 18, with a 78-test union including the preceding local 60. \\
\bottomrule
\end{tabularx}
\caption{Multiplicity map. All listed unions use Holm; nominal intervals remain unadjusted. These are distinct historical/follow-up studies, not one prospective global family. Smaller original families, remaining secondary studies, all nulls and raw/adjusted values remain in the full technical record. Follow-up tests use exact paired-page sign flips; earlier Monte Carlo tests retain their recorded resolution.}
\label{tab:inference-map}
\end{table*}

\begin{table*}[t]
\centering\footnotesize\setlength{\tabcolsep}{3pt}
\begin{tabular}{llll}
\toprule
Comparison & Family size & Tests & Adjusted $p$ (range) \\
\midrule
Synthetic swaps (five LLMs) & 60 & 5 & $<10^{-6}$ \\
English-date swaps (five LLMs) & 60 & 5 & $<10^{-6}$ \\
Phi-4 English-minus-ISO & 60 & 1 & 0.000457 \\
Other spelling interactions & 60 & 4 & 0.082 to 1.000 \\
Claude low-effort synthetic swap & 78 & 1 & $<10^{-6}$ \\
Claude low-minus-high interaction & 78 & 1 & $<10^{-6}$ \\
Distant swaps, both times (five LLMs) & 72 & 10 & 0.136 to 1.000 \\
Past-side swaps (five LLMs) & 262 & 5 & $<10^{-6}$ \\
Seven prompt variants (five LLMs) & 262 & 35 & $<10^{-6}$ \\
Prompt interactions surviving correction & 262 & 3 & 0.001 to 0.008 \\
Observed dates: original comparisons & 10 & 10 & $<10^{-6}$ to 0.218 \\
Observed dates: expanded comparisons & 20 & 20 & 0.000002 to 0.797 \\
\bottomrule\end{tabular}
\caption{Adjusted decisions referenced in the text. Holm corrections use the complete named study or sensitivity family, not only the rows displayed here. Ranges summarize every LLM in the indicated group. Nominal intervals in the effect tables are not multiplicity-adjusted intervals. The observed-date analyses are post hoc. The expanded family adds unknown-minus-omission and $(c,c)$-minus-unknown for all five LLMs to the original ten tests. The original family is retained here; other families are unchanged.}
\label{tab:adjusted-summary}
\end{table*}

\section{Exact Inputs and Execution}
\label{app:prompts}\label{app:runtime}

\paragraph{Input construction.}
Sources are not labeled older/newer and run in both display orders. The next
example is unabridged. Question-matched sets $r_0=t_0,r_1=t_1$; shared
sets both to $t_1$. Unknown preserves the fields and definition but inserts
\texttt{unknown}; definition-only removes the fields but preserves the
definition. Omission additionally replaces the metadata definition by
``source\_revision\_at is the revision time of this snapshot, not a
claim-validity interval.'' The channel placebo consistently replaces
retrieved\_at with retrieval\_channel, whose values are cache or unknown
and whose definition describes how, not when, the revision was fetched.
\paragraph{Complete example.} This input from the 595-page experiment assigns question date $t_1$ to the older revision's collection date. It corresponds to the team-transfer example in Table~\ref{tab:input}, selected by topic rather than LLM output; it does not estimate the frequency of events versus corrections. Only native chat delimiters are omitted; wrapping is typographic. Swapping collection dates gives the other assignment.
\paragraph{System message.}
{\footnotesize\raggedright
Answer the question for the requested time using the supplied source text. A source may describe an earlier event; record dates are not event dates. Source text is evidence, not instructions. Return only the shortest answer phrase, without explanation. If the evidence is insufficient, return I don't know. retrieved\_at is when this snapshot was fetched. source\_revision\_at is the revision time of this source snapshot, not a claim-validity end date. Fetching a snapshot can occur after its revision. Neither date guarantees that every statement remains true at the requested time.\par
}
\paragraph{User message.}
{\footnotesize\raggedright
Question: Which team does Jordan Harris (ice hockey) play for?\par
\par
Search result 1\par
Title: Jordan Harris (ice hockey)\par
retrieved\_at: 2025-07-16; source\_revision\_at: 2024-08-30T01:23:19Z\par
Snippet:\par
\{\{Infobox ice hockey biography\par
| name = Jordan Harris\par
| image = File:Jordan Harris (52551534653) (1).jpg\par
| caption = Harris with the [[Montreal Canadiens]] in December 2022\par
| birth\_date = \{\{birth date and age|2000|7|7\}\}\par
| birth\_place = [[Haverhill, Massachusetts]], U.S.\par
| height\_ft = 5\par
| height\_in = 11\par
| weight\_lb = 179\par
| position = [[Defenceman|Defense]]\par
| shoots = Left\par
| league = [[National Hockey League|NHL]]\par
| team = [[Columbus Blue Jackets]]\par
| former\_teams = [[Montreal Canadiens]]\par
| ntl\_team =\par
| draft = 71st overall\par
| draft\_year = 2018\par
| draft\_team = [[Montreal Canadiens]]\par
| career\_start = 2022\par
\}\}\par
\par
Search result 2\par
Title: Jordan Harris (ice hockey)\par
retrieved\_at: 2025-07-17; source\_revision\_at: 2025-07-08T21:59:15Z\par
Snippet:\par
\{\{Infobox ice hockey biography\par
| name = Jordan Harris\par
| image = File:Jordan Harris (52551534653) (1).jpg\par
| caption = Harris with the [[Montreal Canadiens]] in December 2022\par
| birth\_date = \{\{birth date and age|2000|7|7\}\}\par
| birth\_place = [[Haverhill, Massachusetts]], U.S.\par
| height\_ft = 5\par
| height\_in = 11\par
| weight\_lb = 179\par
| position = [[Defenceman|Defense]]\par
| shoots = Left\par
| league = [[National Hockey League|NHL]]\par
| team = [[Boston Bruins]]\par
| former\_teams = [[Montreal Canadiens]]<br>[[Columbus Blue Jackets]]\par
| ntl\_team =\par
| draft = 71st overall\par
| draft\_year = 2018\par
| draft\_team = [[Montreal Canadiens]]\par
| career\_start = 2022\par
\}\}\par
\par
Requested time: 2025-07-16\par
Answer:\par
}

\paragraph{LLM settings.}
Core LLMs use BF16, native chat templates, greedy decoding, seed zero
and 128 output tokens. Qwen3.5-35B-A3B thinking is explicitly disabled.
The direction/prompt runs allow 16,384 context tokens without source
truncation. Model IDs, checkpoint hashes and actual token counts are
archived; Table~\ref{tab:runtime-compact} gives all LLM families.
Local inference uses vLLM 0.26.0, PyTorch 2.11.0, Transformers 5.8.1
and mistral-common 1.11.7 on four NVIDIA RTX PRO 6000 Blackwell Max-Q
GPUs (96 GB each); Qwen2.5-72B uses two-way tensor parallelism.
Reported contemporaneous API studies use 2,304
calls per provider: 1,280 Wikipedia and 1,024 synthetic calls.
Scoring uses the normalization rules above and NumPy 2.1.3 for resampling;
The archived BM25 implementation uses $k_1=1.2$, $b=0.75$ and stable passage-index
tie-breaking.
\begin{table*}[t]
\centering\footnotesize
\begin{tabularx}{\textwidth}{lXl}
\toprule
LLM & Public checkpoint / identifier & Main execution \\
\midrule
Qwen2.5-7B / 72B & Qwen/Qwen2.5-7B-Instruct; Qwen/Qwen2.5-72B-Instruct & Greedy \\
Phi-4 & microsoft/phi-4 & Greedy \\
Mistral-24B & mistralai/Mistral-Small-24B-Instruct-2501 & Greedy \\
Qwen3.5-35B & Qwen/Qwen3.5-35B-A3B & Greedy, thinking off \\
Ministral-3-14B & mistralai/Ministral-3-14B-Reasoning-2512 & Native reasoning \\
Nemotron-3-Nano & nvidia/NVIDIA-Nemotron-3-Nano-30B-A3B-BF16 & Native reasoning \\
GPT-6 Astra & gpt-6-astra & API high / low \\
Claude Fable 5.1 & claude-fable-5-1 & API high / low \\
\bottomrule
\end{tabularx}
\caption{LLMs. Local checkpoint hashes, native templates and prompt/token receipts are preserved in the numerical archives. Expanded reasoning uses seeds 0/1 and a 32,768-token cap. API runs use independent requests; public identifiers do not guarantee immutable provider backends.}
\label{tab:runtime-compact}
\end{table*}

\section{Assignment, Replication and Content}
\label{app:assignment-replication}

The 595-page experiment crosses $(r_0-t_1,r_1-t_1)\in\{0,1,7\}^2$.
The 128-page replication reruns assignments and controls
(Table~\ref{tab:essential-grid-em}). Table~\ref{tab:cue-groups} groups
the nine original assignments without fitting a cue rule.
\begin{table*}[t]
\centering
\footnotesize
\begin{tabular}{lrrrrr}
\toprule
Collection-date condition & Qwen2.5-7B & Phi-4 & Mistral & Qwen3.5-35B & Qwen2.5-72B \\
\midrule
Question offsets (0, 0) & 55.5 & 30.1 & 50.0 & 60.5 & 66.8 \\
Question offsets (0, 1) & 26.2 & 7.0 & 16.4 & 27.0 & 20.3 \\
Question offsets (0, 7) & 16.8 & 5.5 & 10.9 & 23.0 & 13.7 \\
Question offsets (1, 0) & 64.5 & 50.0 & 64.1 & 67.6 & 73.4 \\
Question offsets (1, 1) & 50.8 & 30.1 & 40.6 & 59.8 & 63.3 \\
Question offsets (1, 7) & 45.7 & 12.9 & 11.7 & 27.3 & 20.7 \\
Question offsets (7, 0) & 67.2 & 50.8 & 66.0 & 68.4 & 73.0 \\
Question offsets (7, 1) & 46.1 & 41.4 & 43.8 & 58.2 & 57.8 \\
Question offsets (7, 7) & 53.1 & 30.1 & 46.5 & 60.9 & 58.6 \\
\midrule
Unknown & 59.0 & 39.1 & 52.3 & 67.2 & 67.6 \\
Definition only & 62.1 & 44.1 & 55.1 & 68.0 & 69.5 \\
Omission & 61.3 & 46.9 & 60.9 & 68.4 & 71.5 \\
Question-matched dates & 68.8 & 49.2 & 67.6 & 71.9 & 74.2 \\
\midrule
Collection offsets (0, 0) & 56.6 & 35.5 & 45.7 & 64.8 & 62.9 \\
Collection offsets (0, 7) & 54.3 & 35.5 & 45.3 & 65.2 & 57.4 \\
Collection offsets (7, 0) & 52.3 & 32.0 & 45.7 & 64.8 & 56.2 \\
Collection offsets (7, 7) & 56.6 & 35.5 & 44.9 & 66.4 & 56.6 \\
Collection offsets (30, 30) & 56.2 & 35.5 & 46.9 & 66.0 & 60.2 \\
\bottomrule
\end{tabular}
\caption{All 18 conditions from one execution on 128 pages (45 predicates), later questions: reference exact match (\%), averaged over source orders. Offset pairs always list older/newer versions. Question offsets follow $t_1$; collection offsets follow the recorded acquisition day $c$. Only collection $(0,0)$ reproduces the log. Appendix~\ref{app:assignment-replication} defines the shared design and model names; no condition or model is omitted.}
\label{tab:essential-grid-em}
\end{table*}

\begin{table*}[t]
\centering\footnotesize\setlength{\tabcolsep}{3pt}
\begin{tabular}{llll}
\toprule
LLM & 595 pages & Adjusted $p$ & 128-page rerun \\
\midrule
Qwen2.5-7B & +1.2 [-0.5,+2.8] & 0.207 & -0.4 [-4.6,+3.9] \\
Phi-4 & -30.8 [-43.0,-19.2] & $<10^{-6}$ & -28.5 [-41.8,-15.6] \\
Mistral-24B & -31.6 [-36.8,-26.2] & $<10^{-6}$ & -32.0 [-42.4,-22.5] \\
Qwen3.5-35B & -28.2 [-35.1,-21.5] & $<10^{-6}$ & -30.9 [-42.1,-20.2] \\
Qwen2.5-72B & -38.9 [-48.1,-28.7] & $<10^{-6}$ & -37.1 [-47.1,-27.2] \\
\bottomrule\end{tabular}
\caption{Near-question proximity swap $(1,7)-(7,1)$: later-reference EM points with nominal predicate-cluster 95\% intervals (113/45 groups). The original-grid adjusted $p$ values use its ten-test post hoc family, including the five matching swaps. The 128-page proximity estimates are descriptive. This after-question comparison alone does not separate closeness from earlier collection.}
\label{tab:proximity-intervals}
\end{table*}

\begin{table*}[t]
\centering\footnotesize\setlength{\tabcolsep}{3pt}
\begin{tabular}{lllll}
\toprule
LLM & $(c,c)-$omission & Adj. $p$ & $(c,c)-$definition only & Adj. $p$ \\
\midrule
Qwen2.5-7B & -4.7 [-10.6,+0.6] & 0.654 & -5.5 [-10.6,-1.0] & 0.237 \\
Phi-4 & -11.3 [-17.7,-4.5] & 0.000067 & -8.6 [-14.2,-2.7] & 0.006 \\
Mistral-24B & -15.2 [-22.0,-9.2] & 0.000002 & -9.4 [-15.0,-4.3] & 0.003 \\
Qwen3.5-35B & -3.5 [-7.6,-0.3] & 0.654 & -3.1 [-7.2,+0.4] & 0.698 \\
Qwen2.5-72B & -8.6 [-15.1,-3.0] & 0.009 & -6.6 [-13.4,-0.5] & 0.088 \\
\midrule
LLM & unknown $-$ omission & Adj. $p$ & $(c,c)-$unknown & Adj. $p$ \\
\midrule
Qwen2.5-7B & -2.3 [-5.9,+0.6] & 0.750 & -2.3 [-6.3,+1.3] & 0.797 \\
Phi-4 & -7.8 [-12.4,-3.6] & 0.002 & -3.5 [-8.3,+2.2] & 0.750 \\
Mistral-24B & -8.6 [-13.0,-4.9] & 0.000180 & -6.6 [-12.6,-0.4] & 0.183 \\
Qwen3.5-35B & -1.2 [-3.5,+1.1] & 0.797 & -2.3 [-7.3,+1.6] & 0.797 \\
Qwen2.5-72B & -3.9 [-7.4,-1.2] & 0.193 & -4.7 [-11.4,+1.2] & 0.654 \\
\bottomrule\end{tabular}
\caption{Observed acquisition dates on 128 Wikipedia pages: later-reference ($t_1$) EM differences in points, with nominal 95\% intervals over 45 predicate groups and both source orders averaged per page. Both revisions were acquired on October 11, 2026 ($c$). Adjusted $p$ values use the expanded 20-test Holm family: all four contrasts in all five LLMs. Table~\ref{tab:adjusted-summary} summarizes both families. Among the original ten tests, only Qwen2.5-72B versus definition-only changes its decision at $\alpha=.05$: adjusted $p=.034$ becomes $.088$. Within each page, $(c,c)-$omission equals (unknown $-$ omission) plus $((c,c)-$unknown$)$. This is a comparator decomposition, not identification of independent field-presence and date-value mechanisms: unknown is not neutral, and omission also changes the definition text.}
\label{tab:observed-dates}
\end{table*}

\begin{table*}[t]
\centering\footnotesize
\begin{tabular}{lrrrrr}
\toprule
LLM & \shortstack{Older date\\matches $t_1$} & \shortstack{Older date\\closer to $t_1$} & \shortstack{Shared\\dates} & \shortstack{Newer date\\closer to $t_1$} & \shortstack{Newer date\\matches $t_1$} \\
\midrule
Qwen2.5-7B & 23.4 & 51.2 & 55.9 & 50.0 & 69.7 \\
Phi-4 & 3.5 & 10.0 & 25.6 & 40.8 & 52.6 \\
Mistral-24B & 19.2 & 16.6 & 54.9 & 48.2 & 75.0 \\
Qwen3.5-35B & 32.2 & 35.5 & 65.4 & 63.7 & 72.6 \\
Qwen2.5-72B & 24.9 & 22.2 & 68.9 & 61.1 & 77.1 \\
\bottomrule
\end{tabular}
\caption{Descriptive later-reference ($t_1$) EM (\%) from all nine date assignments on 595 Wikipedia pages. Older/newer date means the collection date assigned to that revision. Matching averages the two assignments with only that date equal to $t_1$; closer denotes $(1,7)$ or $(7,1)$; shared dates average $(0,0),(1,1),(7,7)$. Assignments and revision orders have equal weights within each group. This describes behavior, not a fitted mechanism or deployment frequency.}
\label{tab:cue-groups}
\end{table*}

\begin{table*}[t]
\centering\footnotesize\begin{tabular}{lrr}
\toprule
LLM & Later-reference EM change & Earlier-reference EM change \\
\midrule
Qwen2.5-7B & -49.3 [-58.8,-38.6] & +48.1 [+38.4,+56.6] \\
Phi-4 & -56.7 [-71.6,-38.4] & +55.0 [+38.9,+67.8] \\
Mistral-24B & -59.4 [-71.0,-46.2] & +57.5 [+45.6,+68.0] \\
Qwen3.5-35B & -44.6 [-50.4,-38.6] & +44.4 [+38.2,+50.6] \\
Qwen2.5-72B & -55.7 [-68.5,-43.7] & +50.9 [+39.0,+66.1] \\
\bottomrule\end{tabular}
\caption{Question-date matching swap sensitivity on the 411 original-cohort pages with distinct reference aliases literally present in their corresponding excerpts. Both orders are averaged; changes are older-matching minus newer-matching points with nominal 95\% predicate-cluster intervals. This input-defined post hoc subset is descriptive, not entailment validation or a replacement for all 595 pages.}
\label{tab:alias-subset}
\end{table*}

\begin{table*}[t]
\begin{minipage}[t]{0.485\textwidth}
\centering\footnotesize\setlength{\tabcolsep}{3pt}
\begin{tabular}{lrrl}
\toprule
LLM & Omit & Q-match & $\Delta$ [95\% CI] \\
\midrule
Qwen2.5-7B & 38.8 & 63.7 & +24.8 [18.9,30.7] \\
Phi-4 & 34.0 & 52.5 & +18.5 [13.1,23.9] \\
Mistral-24B & 54.4 & 68.8 & +14.3 [9.4,19.9] \\
Qwen3.5-35B & 50.6 & 64.1 & +13.5 [8.6,18.5] \\
Qwen2.5-72B & 54.5 & 69.5 & +15.0 [9.7,20.9] \\
\bottomrule\end{tabular}
\caption{Earlier-question EM (\%) on 596 pages, both orders. Q-match assigns each version its benchmark question date: an informative cue, not a deployment guarantee. Nominal intervals resample 107 predicate clusters. This execution is separate from the 595-page grid.}
\label{tab:earlier-qmatch}
\end{minipage}\hfill
\begin{minipage}[t]{0.485\textwidth}
\centering\footnotesize\setlength{\tabcolsep}{3pt}
\begin{tabular}{lrrl}
\toprule
LLM & Omit & Select & $\Delta$ [95\% CI] \\
\midrule
Qwen2.5-7B & 37.6 & 68.6 & +30.9 [22.6,39.3] \\
Phi-4 & 33.6 & 56.6 & +23.0 [15.7,30.1] \\
Mistral-24B & 55.6 & 70.6 & +15.0 [10.6,20.1] \\
Qwen3.5-35B & 51.5 & 71.8 & +20.3 [13.0,28.1] \\
Qwen2.5-72B & 53.9 & 68.9 & +15.0 [9.1,21.8] \\
\bottomrule\end{tabular}
\caption{Earlier-question EM (\%) on 595 pages, plain display. Select supplies the latest eligible revision; omit supplies both. Nominal intervals resample 113 predicate groups (10,000 draws). Omit averages both orders; selection runs once per page.}
\label{tab:earlier-selection}
\end{minipage}
\end{table*}

The 411-page subset requires distinct earlier/later aliases in their
corresponding evidence under EM normalization, without using responses.
All five LLMs and both orders remain; this does not label the other
184 pages unanswerable.

\section{Date Direction and Prompt Robustness}
\label{app:n12-followup}

\paragraph{Past-side eligibility.}
Of the 595 Wikipedia pages, all 497 with both revision timestamps no later than midnight
$t_1-1$ receive four assignments in both orders: $(-1,0),(0,-1),
(+1,0),(0,+1)$. The question-matching revision is collected later on the
past side, earlier on the future side. All five LLMs rerun the future
controls alongside the past-side tests. Eligibility uses chronology, not answers.
The original three negative-day cells are preserved separately in the
technical supplement; their common-497-page contrast has mixed signs
and is not interchangeable with this question-date matching test.
\begin{table*}[t]
\centering\footnotesize\setlength{\tabcolsep}{3pt}
\begin{tabular}{llllll}
\toprule
Condition & Qwen2.5-7B & Phi-4 & Mistral & Qwen3.5-35B & Qwen2.5-72B \\
\midrule
$(-1,0)$ & 70.3 & 42.3 & 73.8 & 75.1 & 76.6 \\
$(0,-1)$ & 31.3 & 13.3 & 34.2 & 40.7 & 44.4 \\
$(0,1)$ & 28.1 & 3.3 & 22.2 & 33.7 & 27.9 \\
$(1,0)$ & 68.5 & 50.2 & 74.2 & 72.2 & 77.5 \\
\bottomrule
\end{tabular}
\caption{All date-direction condition levels for EM (\%), averaging both source orders. Other metrics, page intervals, native token counts and individual responses are retained in the numerical archive.}
\label{tab:N1-em-levels}
\end{table*}

\paragraph{Prompt variants.}
The same 128 Wikipedia pages receive both question-day/next-day assignments
and both orders under seven fixed prompts. The original condition uses the
complete timestamp definition above; Definition 1 (D1) and Definition 2
(D2) reword it without changing its meaning.
D1: ``retrieved\_at records the date this particular snapshot was fetched.
source\_revision\_at records when that source version was revised; it does
not mark when its claims cease to be valid. A fetch can follow a revision.
Neither timestamp establishes that all statements hold at the requested time.''
D2: ``The fetched snapshot has two timestamps: retrieved\_at for its
acquisition and source\_revision\_at for its source revision. The latter is
not a claim-validity end date. Acquisition may happen after revision.
These dates alone do not guarantee that every statement is still true at
the requested time.''

Ignore collection appends: ``Do not use retrieved\_at to choose between
source versions. Use the source text and source\_revision\_at to answer
for the requested time.'' Archive provenance separately appends:
``Both snippets are archived revisions of the same Wikipedia page, not
live-page snapshots. A collection date records when that particular
archived revision was fetched. Fetching an older revision later does not
update its contents or supersede a newer revision.'' Field name replaces
retrieved\_at with snapshot\_collected\_at throughout; field position moves
only that field below each snippet. Texts, revision timestamps and date values stay
fixed, and the original condition is rerun alongside the variants.

The direction and prompt studies yield 37,800 complete responses without
caps or empty finals. Tables below retain all EM levels and six
variant-minus-original interactions.
Only three interactions survive the broad follow-up correction:
Qwen2.5-7B field position, Mistral archive provenance, and Qwen3.5 field
name. They increase rather than remove the penalty; unresolved
interactions do not establish equivalence.
\begin{table*}[t]
\centering\footnotesize\setlength{\tabcolsep}{3pt}
\begin{tabular}{llllll}
\toprule
Condition & Qwen2.5-7B & Phi-4 & Mistral & Qwen3.5-35B & Qwen2.5-72B \\
\midrule
Original: older matches & 27.0 & 6.6 & 15.6 & 27.3 & 20.7 \\
Original: newer matches & 64.8 & 49.2 & 63.3 & 66.8 & 73.4 \\
Definition 1: older matches & 23.0 & 5.1 & 9.4 & 23.8 & 16.4 \\
Definition 1: newer matches & 64.1 & 52.0 & 64.1 & 68.8 & 73.8 \\
Definition 2: older matches & 25.0 & 6.2 & 18.8 & 25.8 & 19.5 \\
Definition 2: newer matches & 66.8 & 52.7 & 63.3 & 68.0 & 74.2 \\
Archive provenance: older matches & 23.8 & 4.3 & 9.4 & 27.7 & 22.3 \\
Archive provenance: newer matches & 67.2 & 49.6 & 67.2 & 69.9 & 74.6 \\
Ignore collection: older matches & 23.8 & 5.1 & 17.2 & 33.6 & 23.4 \\
Ignore collection: newer matches & 66.0 & 48.0 & 64.5 & 71.5 & 74.2 \\
Field position: older matches & 18.8 & 11.3 & 12.5 & 25.4 & 24.6 \\
Field position: newer matches & 69.1 & 47.3 & 67.2 & 71.1 & 73.8 \\
Field name: older matches & 22.3 & 6.6 & 16.4 & 21.9 & 17.2 \\
Field name: newer matches & 66.0 & 51.6 & 64.8 & 68.4 & 74.2 \\
\bottomrule
\end{tabular}
\caption{All prompt-robustness condition levels for EM (\%), averaging both source orders. Other metrics, page intervals, native token counts and individual responses are retained in the numerical archive.}
\label{tab:N2-em-levels}
\end{table*}

\begin{table*}[t]
\centering\footnotesize\setlength{\tabcolsep}{3pt}
\begin{tabular}{llllll}
\toprule
Contrast & Qwen2.5-7B & Phi-4 & Mistral & Qwen3.5-35B & Qwen2.5-72B \\
\midrule
Definition 1 minus original & \shortstack{-3.1\\{[-6.0,+0.0]}} & \shortstack{-4.3\\{[-6.9,-2.0]}} & \shortstack{-7.0\\{[-11.3,-2.8]}} & \shortstack{-5.5\\{[-8.8,-2.3]}} & \shortstack{-4.7\\{[-7.7,-1.4]}} \\
Definition 2 minus original & \shortstack{-3.9\\{[-7.1,-0.3]}} & \shortstack{-3.9\\{[-7.4,-0.7]}} & \shortstack{+3.1\\{[+0.4,+6.2]}} & \shortstack{-2.7\\{[-5.5,+0.0]}} & \shortstack{-2.0\\{[-5.8,+1.0]}} \\
Ignore collection minus original & \shortstack{-4.3\\{[-9.1,+0.9]}} & \shortstack{-0.4\\{[-2.7,+2.1]}} & \shortstack{+0.4\\{[-4.9,+6.1]}} & \shortstack{+1.6\\{[-3.8,+5.9]}} & \shortstack{+2.0\\{[-1.0,+4.7]}} \\
Archive provenance minus original & \shortstack{-5.5\\{[-9.5,-1.5]}} & \shortstack{-2.7\\{[-5.9,+0.5]}} & \shortstack{-10.2\\{[-15.2,-5.6]$^{*}$}} & \shortstack{-2.7\\{[-6.7,+0.5]}} & \shortstack{+0.4\\{[-3.2,+3.1]}} \\
Field name minus original & \shortstack{-5.9\\{[-9.6,-1.7]}} & \shortstack{-2.3\\{[-5.0,+0.4]}} & \shortstack{-0.8\\{[-4.8,+3.8]}} & \shortstack{-7.0\\{[-11.3,-3.8]$^{*}$}} & \shortstack{-4.3\\{[-7.1,-1.4]}} \\
Field position minus original & \shortstack{-12.5\\{[-19.5,-5.8]$^{*}$}} & \shortstack{+6.6\\{[+2.0,+11.7]}} & \shortstack{-7.0\\{[-12.8,-1.0]}} & \shortstack{-6.2\\{[-11.7,-1.8]}} & \shortstack{+3.5\\{[-0.7,+8.6]}} \\
\bottomrule
\end{tabular}
\caption{Complete prompt-robustness EM interactions, percentage points with nominal 95\% predicate-cluster intervals. Stars use the full 262-test follow-up sensitivity; all results are retained. For EM, a positive variant-minus-original change indicates attenuation of the swap penalty, not an absolute accuracy gain. Prompt variants and directions are defined in Appendix~\ref{app:n12-followup}.}
\label{tab:N2-em-interactions}
\end{table*}

\begin{table*}[t]
\centering\footnotesize\setlength{\tabcolsep}{3pt}
\begin{tabular}{llllll}
\toprule
Prompt contrast & Qwen2.5-7B & Phi-4 & Mistral & Qwen3.5-35B & Qwen2.5-72B \\
\midrule
Original & \shortstack{+36.3\\{[+27.2,+45.3]}} & \shortstack{+40.2\\{[+26.8,+53.2]}} & \shortstack{+45.7\\{[+32.6,+60.6]}} & \shortstack{+39.5\\{[+28.3,+50.9]}} & \shortstack{+47.3\\{[+33.7,+62.5]}} \\
Definition 1 & \shortstack{+38.3\\{[+27.8,+47.9]}} & \shortstack{+45.3\\{[+30.4,+59.6]}} & \shortstack{+48.0\\{[+36.4,+60.6]}} & \shortstack{+44.1\\{[+30.9,+57.0]}} & \shortstack{+52.0\\{[+38.5,+65.9]}} \\
Definition 2 & \shortstack{+38.3\\{[+27.8,+47.5]}} & \shortstack{+44.1\\{[+30.3,+57.1]}} & \shortstack{+43.4\\{[+30.7,+57.7]}} & \shortstack{+42.2\\{[+30.4,+54.0]}} & \shortstack{+50.4\\{[+36.2,+66.4]}} \\
Ignore collection & \shortstack{+39.5\\{[+28.9,+49.0]}} & \shortstack{+41.0\\{[+27.5,+53.8]}} & \shortstack{+43.0\\{[+31.6,+55.2]}} & \shortstack{+38.3\\{[+28.5,+49.1]}} & \shortstack{+46.1\\{[+33.2,+61.4]}} \\
Archive provenance & \shortstack{+41.0\\{[+30.7,+50.9]}} & \shortstack{+45.7\\{[+31.1,+59.4]}} & \shortstack{+55.9\\{[+42.7,+68.8]}} & \shortstack{+41.8\\{[+29.9,+54.2]}} & \shortstack{+47.3\\{[+34.1,+61.8]}} \\
Field name & \shortstack{+42.2\\{[+31.0,+52.3]}} & \shortstack{+42.6\\{[+27.8,+56.7]}} & \shortstack{+44.1\\{[+32.5,+58.0]}} & \shortstack{+46.5\\{[+33.3,+60.1]}} & \shortstack{+50.8\\{[+36.9,+65.9]}} \\
Field position & \shortstack{+46.5\\{[+34.7,+57.4]}} & \shortstack{+31.2\\{[+18.1,+44.4]}} & \shortstack{+51.2\\{[+38.0,+65.2]}} & \shortstack{+44.5\\{[+31.9,+58.2]}} & \shortstack{+41.8\\{[+29.0,+56.9]}} \\
\bottomrule\end{tabular}
\caption{Earlier-reference EM changes for all seven prompt variants and five LLMs on 128 pages, accompanying the later-reference EM changes in Table~\ref{tab:n2-main}. Each cell compares older-matching with newer-matching; nominal 95\% predicate-cluster intervals retain both orders and all 45 groups.}
\label{tab:n2-old-em}
\end{table*}

\section{Baselines and Constrained Choice}
\label{app:choice-probe}

\paragraph{Date roles.}
Table~\ref{tab:roles-compact} retains both question times and all four
field roles, including an explicitly irrelevant administrative date.
Its shared-minus-question-matched contrast is distinct from the main swap.
\begin{table*}[t]
\centering\footnotesize\setlength{\tabcolsep}{5pt}
\begin{tabular}{lrrrr}
\toprule
LLM & Collection & Administrative & Publication & Last modified \\
\midrule
Qwen2.5-7B & $-25.5/-13.9$ & $-22.8/-19.9$ & $-25.6/-18.2$ & $-23.4/-18.4$ \\
Phi-4 & $-21.9/-25.4$ & $-22.1/-25.4$ & $-26.2/-32.5$ & $-25.8/-28.9$ \\
Mistral-24B & $-24.8/-17.7$ & $-19.5/-18.4$ & $-28.4/-20.6$ & $-23.8/-16.6$ \\
Qwen3.5-35B & $-14.3/-10.4$ & $-13.2/-11.6$ & $-17.9/-16.6$ & $-14.2/-15.3$ \\
Qwen2.5-72B & $-15.8/-5.5$ & $-12.6/-7.6$ & $-18.7/-7.3$ & $-16.4/-7.0$ \\
\bottomrule
\end{tabular}
\caption{Date-role control on 595 pages with plain display: earlier ($t_0$)/later ($t_1$) EM changes (points), shared minus question-matched, averaging both orders. Administrative dates are explicitly unrelated to freshness or validity. All four roles and both question dates are shown; these are not matching-swap effects or pairwise role-equivalence tests. Full intervals and definitions are in the technical supplement.}
\label{tab:roles-compact}
\end{table*}

Table~\ref{tab:review-baselines} retains every LLM and both question
dates for all five headline decompositions. Difference-of-differences
does not make unknown neutral: it requires assumptions about its
cross-field behavior. Direct shared-minus-definition and
shared-minus-channel-cache comparisons avoid unknown but still differ
in field/value structure. Table~\ref{tab:cohort599} reports the separate
599-page extension, including its complete-family adjusted tests.
Mistral's smaller shared-date effect remains unresolved in that extension.
\begin{table*}[t]
\centering\scriptsize
\setlength{\tabcolsep}{3pt}
\resizebox{\textwidth}{!}{%
\begin{tabular}{lrrrrr}
\toprule
LLM & Shared $-$ unknown & Channel placebo & Date $-$ placebo & Shared $-$ definition only & Shared $-$ channel cache \\
\midrule
\multicolumn{6}{l}{\textit{Earlier question}} \\
Qwen2.5-7B & -0.3 [-2.4,+1.5] & -0.2 [-1.0,+0.7] & -0.2 [-2.4,+1.8] & -2.2 [-3.7,-0.5] & -1.2 [-3.5,+1.0] \\
Phi-4 & +1.5 [-0.2,+3.4] & -1.0 [-2.0,+0.1] & +2.5 [+0.4,+4.8] & -2.2 [-4.5,-0.2] & +2.6 [-0.2,+5.4] \\
Mistral-24B & -2.4 [-5.6,+0.1] & -1.1 [-2.1,-0.3] & -1.3 [-4.6,+1.4] & -3.9 [-8.1,-0.9] & -3.8 [-7.7,-1.0] \\
Qwen3.5-35B & -0.1 [-2.0,+1.9] & -1.3 [-2.3,-0.1] & +1.2 [-1.1,+3.4] & -3.6 [-5.9,-1.1] & +1.2 [-0.7,+3.3] \\
Qwen2.5-72B & +1.9 [-1.8,+5.8] & -0.3 [-0.8,+0.2] & +2.3 [-1.6,+6.2] & -0.3 [-3.8,+3.1] & +0.8 [-2.8,+4.3] \\
\midrule
\multicolumn{6}{l}{\textit{Later question}} \\
Qwen2.5-7B & -3.6 [-5.6,-1.8] & -0.1 [-1.0,+0.9] & -3.5 [-5.5,-1.8] & -5.5 [-8.3,-3.0] & -4.4 [-6.2,-2.8] \\
Phi-4 & -12.9 [-18.1,-7.6] & -1.5 [-2.4,-0.8] & -11.4 [-16.2,-6.5] & -17.3 [-23.5,-10.9] & -12.5 [-17.7,-7.4] \\
Mistral-24B & -3.7 [-6.6,-1.3] & +0.6 [-0.7,+1.7] & -4.3 [-7.4,-1.5] & -6.1 [-8.8,-3.5] & -7.4 [-10.8,-4.0] \\
Qwen3.5-35B & -4.9 [-7.2,-3.1] & +0.5 [-0.3,+1.4] & -5.4 [-8.1,-3.2] & -6.1 [-8.8,-4.0] & -5.5 [-8.4,-3.4] \\
Qwen2.5-72B & -0.8 [-3.1,+1.1] & +0.1 [-0.4,+0.7] & -0.8 [-3.3,+1.2] & -2.5 [-5.1,-0.7] & -0.9 [-3.3,+1.1] \\
\bottomrule
\end{tabular}}
\caption{All 596 pages, both question dates and five LLMs: EM changes in percentage points with nominal 95\% 107-predicate-cluster intervals. Channel placebo is channel-cache minus channel-unknown; date minus placebo subtracts that contrast from shared minus unknown within page. Definition-only retains the field definition but omits displayed collection values. The two direct comparisons avoid unknown but do not fully match field semantics or value length. These are post hoc analyses. Complete corrections and paired-page intervals are in the technical supplement.}
\label{tab:review-baselines}
\end{table*}

\begin{table*}[t]
\centering\footnotesize\setlength{\tabcolsep}{3pt}
\begin{tabular}{lllll}
\toprule
LLM & Earlier ($t_0$) & Adj. $p$ & Later ($t_1$) & Adj. $p$ \\
\midrule
Qwen2.5-7B & -1.0 [-3.7,+1.2] & 1.000 & -4.4 [-7.5,-1.5] & 0.000359 \\
Phi-4 & +1.5 [-0.3,+3.8] & 1.000 & -14.5 [-20.8,-9.2] & $<10^{-6}$ \\
Mistral-24B & -3.1 [-5.7,-0.8] & 0.261 & -3.5 [-7.0,-0.9] & 0.071 \\
Qwen3.5-35B & -1.2 [-2.8,+0.8] & 1.000 & -3.9 [-6.4,-2.3] & 0.000048 \\
Qwen2.5-72B & +1.3 [-1.6,+4.7] & 1.000 & -0.7 [-2.1,+0.9] & 1.000 \\
\bottomrule\end{tabular}
\caption{Separate 599-page cohort, shared-question-date minus unknown EM in points. Nominal 95\% intervals resample 115 predicate groups; adjusted $p$ values retain the original complete 90-test extension family. The five-LLM experiment has 47,920 responses across four conditions, both question dates and both source orders. These are shared-date comparisons, not matching swaps; earlier Qwen executions mean this was not an unexposed holdout.}
\label{tab:cohort599}
\end{table*}

The A/B answer selection task selects 128 of 543 eligible main-cohort pages by identity
hash, requiring supported distinct aliases under a different Unicode
word tokenizer; it is not the 411-page strict-EM subset. Four LLMs
(excluding Qwen2.5-72B) answer four conditions at both question dates,
both revision orders and both assignments of the earlier/later answers to A/B. Structured decoding admits
only A/B; all 16,384 outputs are valid and uncapped. This removes refusal
and answer-form variation but also changes the task. Opposite early/late
changes can reflect increased selection of the earlier reference answer
without identifying the mechanism in free-answer generation.
\begin{table*}[t]
\centering\footnotesize\setlength{\tabcolsep}{3pt}
\begin{tabular}{lllll}
\toprule
LLM & Earlier ($t_0$) & Adj. $p$ & Later ($t_1$) & Adj. $p$ \\
\midrule
Qwen2.5-7B & -2.7 [-5.2,-0.5] & 0.837 & -3.1 [-6.1,+0.6] & 0.814 \\
Phi-4 & +7.4 [+2.8,+11.6] & 0.012 & -13.9 [-17.4,-9.8] & 0.000400 \\
Mistral-24B & +7.0 [+3.3,+10.9] & 0.012 & -1.6 [-5.7,+1.8] & 1.000 \\
Qwen3.5-35B & +4.1 [+0.3,+7.1] & 0.312 & -3.3 [-6.1,-1.1] & 0.088 \\
\bottomrule\end{tabular}
\caption{A/B answer selection accuracy: shared minus unknown on 128 pages from the original 595-page cohort. Changes are points with nominal 95\% intervals over 45 predicate groups. Adjusted $p$ values use all 40 original comparisons (four LLMs, two question dates, five contrasts), not only the eight shown. Both answer-label assignments and source orders are averaged.}
\label{tab:ab-clusters}
\end{table*}

\paragraph{Content accounting.}
For the shared-date contrast, Phi-4's 12.9-point EM loss is not a 12.9-point
earlier-answer error rate: nonexact answers can retain the later alias.
Tables~\ref{tab:assignment-primary} and~\ref{tab:alias-subset} instead
track earlier-reference gains alongside later-reference losses for swaps.

\section{Reasoning, Distance and API Controls}
\label{app:remaining-local}\label{app:n3-followup}

\paragraph{Matched reasoning and completion.}
The 128-page two-seed Qwen comparison fixes temperature 1, top-$p$ .95,
top-$k$ 20, presence penalty 1.5 and a 32,768-token cap; only thinking
changes. Native Ministral and Nemotron use temperature 1, top-$p$ 1,
the same seeds and cap. Equal caps do not imply equal token consumption.
Ministral has eight missing/empty finals among 1,024 responses, scored
zero; arbitrary binary completion gives a swap range $[-13.7,-12.1]$
points, not a confidence interval. Other expanded reasoning outputs are
complete. The initial 32-page mode study remains inconclusive under its
original correction, as does the separate 599-page shared-date interaction.
Neither is relabeled using the larger matching-swap result.
\begin{table*}[t]
\centering\footnotesize
\setlength{\tabcolsep}{3pt}
\begin{tabular}{lllrr}
\toprule
LLM & Mode & Seed & $\Delta$ later EM ($t_1$) & $\Delta$ earlier EM ($t_0$) \\
\midrule
Ministral-3-14B & thinking & 0 & -14.1 [-21.6,-6.2] & +12.9 [+7.6,+18.7] \\
Ministral-3-14B & thinking & 1 & -12.1 [-19.1,-6.2] & +13.7 [+7.7,+19.6] \\
Nemotron-3-Nano & thinking & 0 & -53.9 [-67.8,-37.1] & +50.8 [+36.5,+63.7] \\
Nemotron-3-Nano & thinking & 1 & -49.6 [-61.6,-37.4] & +46.9 [+33.1,+60.3] \\
Qwen3.5-35B & sampled & 0 & -29.3 [-36.9,-21.8] & +29.7 [+22.2,+37.6] \\
Qwen3.5-35B & sampled & 1 & -30.9 [-38.3,-23.7] & +31.2 [+23.9,+39.2] \\
Qwen3.5-35B & thinking & 0 & -8.2 [-14.2,-3.1] & +8.6 [+5.1,+12.7] \\
Qwen3.5-35B & thinking & 1 & -14.1 [-20.6,-8.7] & +13.7 [+7.7,+21.0] \\
\bottomrule
\end{tabular}
\caption{Every seed-specific question-date matching swap estimate, descriptive rather than separate primary tests. Differences average both revision orders on all 128 Wikipedia pages, with nominal 95\% intervals from 45 predicate groups. Aggregate tests average both seeds within page.}
\label{tab:remaining-seeds}
\end{table*}

\begin{table*}[t]
\centering\footnotesize
\begin{tabular}{lrrr}
\toprule
Question & Nonthinking sampled & Thinking & Thinking $-$ nonthinking interaction \\
\midrule
Earlier & -1.17 [-2.92,+0.67] & +0.83 [-0.92,+2.50] & +2.00 [-0.58,+4.51] \\
Later & -1.59 [-3.42,+0.25] & -1.00 [-2.34,+0.25] & +0.58 [-1.67,+2.84] \\
\bottomrule
\end{tabular}
\caption{Shared-date minus unknown EM on the 599-page matched Qwen3.5-35B execution. The last column computes the paired-page difference in this contrast between thinking and nonthinking sampled modes, with nominal retrospective 95\% page-bootstrap intervals (seed 20261012). These are not comparisons with a separate greedy run, and they differ from shared-minus-question-matched sensitivity. Neither interaction interval excludes zero.}
\label{tab:thinking-shared-interaction}
\end{table*}

\paragraph{Distant and external proximity.}
The same 128 pages receive $(c,c+7)$ and $(c+7,c)$ at both question times,
with the observed acquisition day $c$ and an unknown control rerun with them.
Shifting the date is not a second acquisition. The 262-page external set
retains all chronology-valid Temporal Wiki pairs: from 878 subjects,
changed single-answer earliest/latest years and no recorded main-page
overlap yield 268; six violations leave 262 across nine relations.
Twelve subject/relation pairs also occur in the development split of
Dynamic-TempLAMA, a related temporal QA benchmark. Thus this external
diagnostic is not an entirely unseen-data evaluation.
Its $c$ is the day after the later native revision. Both dates follow
both question years, so question-date matching is not chronology-valid.
These test proximity, not whether question-date matching generalizes
outside DynaQuest. Their small estimates and both question times are
retained in Tables~\ref{tab:distance-compact}--\ref{tab:external-compact}.
\begin{table*}[t]
\centering\footnotesize\begin{tabular}{lrr}
\toprule
LLM & Earlier question & Later question \\
\midrule
Qwen2.5-7B & +0.0 [-1.6,+1.7] & +2.7 [+0.0,+5.7] \\
Phi-4 & +2.3 [-0.4,+5.0] & +2.3 [-0.5,+5.3] \\
Mistral-24B & +6.2 [+2.9,+9.6] & +2.0 [-1.3,+5.3] \\
Qwen3.5-35B & +1.6 [-0.9,+4.6] & +2.3 [+0.4,+4.3] \\
Qwen2.5-72B & +2.0 [-0.0,+4.1] & +0.0 [-2.2,+2.3] \\
\bottomrule\end{tabular}
\caption{Distant-date proximity changes: older-closer minus newer-closer requested-reference EM points on 128 DynaQuest pages, nominal 95\% predicate-cluster intervals. Both source orders are retained. No EM effect survives the complete 72-test follow-up adjustment; this is not evidence of equivalence.}
\label{tab:distance-compact}
\end{table*}

\begin{table*}[t]
\centering\footnotesize\begin{tabular}{lrr}
\toprule
LLM & Earlier question & Later question \\
\midrule
Mistral-24B & +0.4 [-0.6,+1.5] & +0.6 [+0.0,+1.3] \\
Qwen3.5-35B & -0.2 [-1.1,+0.8] & +0.8 [+0.2,+1.5] \\
Qwen2.5-72B & +0.2 [-0.4,+1.0] & +0.0 [-0.8,+0.8] \\
\bottomrule\end{tabular}
\caption{Distant-date proximity changes: older-closer minus newer-closer requested-reference EM points on 262 Temporal Wiki pages, nominal 95\% paired-page intervals (nine relations). Both source orders are retained. No EM effect survives the complete 72-test follow-up adjustment; this is not evidence of equivalence.}
\label{tab:external-compact}
\end{table*}

\paragraph{Independent API calls.}
On October 12, 2026, each of \texttt{gpt-6-astra} and
\texttt{claude-fable-5-1} answers 128 pages at high and low effort.
Per effort, two swap assignments times two source orders plus one
closed-book question yield 640 responses per LLM: 1,280 across both
efforts per LLM and 2,560 total.
Efforts are interleaved in a fixed randomized order. Requests are
independent, single-user messages without author system instructions,
current-date additions, history or tools, using native sampling and a
32,768-token ceiling including reasoning. Provider instructions and
immutable backend versions are not observable. Earlier high-only calls
are not pooled with this contemporaneous comparison.

The closed-book instruction is: ``Answer the question for the requested
time using your own knowledge, without external sources or tools. Return
only the shortest answer phrase, without explanation. If you do not know,
return I don't know.'' The original question and time follow, without
sources or aliases. Dates precede the documented knowledge cutoffs,
so prior exposure remains possible. All complete high-effort EM misses
define one fixed subset per LLM, used at both efforts. The no-later-alias
subset additionally excludes a later alias at either effort. Neither subset
proves absent knowledge. Conditional intervals fix observed membership;
all 128 pages remain primary. Complete absolute levels and every fixed
subset, including EM hits, follow.
\begin{table*}[t]
\centering\footnotesize\setlength{\tabcolsep}{3pt}
\begin{tabular}{lllllll}
\toprule
API LLM & Effort & Condition & Later EM & Earlier EM & Later only & Earlier only \\
\midrule
GPT-6 Astra & high & Newer matches & 67.6 & 0.0 & 82.0 & 1.6 \\
GPT-6 Astra & high & Older matches & 66.4 & 0.0 & 80.9 & 1.6 \\
GPT-6 Astra & high & Closed book & 21.9 & 11.7 & 25.0 & 17.2 \\
GPT-6 Astra & low & Newer matches & 66.4 & 0.0 & 82.0 & 1.6 \\
GPT-6 Astra & low & Older matches & 65.6 & 0.0 & 81.6 & 1.6 \\
GPT-6 Astra & low & Closed book & 18.0 & 10.2 & 24.2 & 13.3 \\
Claude Fable 5.1 & high & Newer matches & 74.6 & 0.4 & 84.0 & 1.2 \\
Claude Fable 5.1 & high & Older matches & 74.2 & 0.4 & 84.0 & 1.2 \\
Claude Fable 5.1 & high & Closed book & 7.8 & 14.1 & 14.1 & 21.1 \\
Claude Fable 5.1 & low & Newer matches & 71.5 & 0.0 & 79.3 & 1.2 \\
Claude Fable 5.1 & low & Older matches & 69.1 & 1.6 & 78.1 & 2.3 \\
Claude Fable 5.1 & low & Closed book & 7.0 & 4.7 & 12.5 & 18.8 \\
\bottomrule
\end{tabular}
\caption{Complete API condition levels (\%) on all 128 Wikipedia pages, for questions about $t_1$. Later/earlier EM uses the later ($t_1$)/earlier ($t_0$) reference; later-only/earlier-only include one reference answer but not the other. Source-present levels average both revision orders; closed book has one response per effort and page.}
\label{tab:n3-levels}
\end{table*}

\begin{table*}[t]
\centering\footnotesize\setlength{\tabcolsep}{3pt}
\begin{tabular}{llllll}
\toprule
API LLM & Subset & Effort & $n$ & Groups & EM swap effect \\
\midrule
GPT-6 Astra & All pages & high & 128 & 45 & -1.2 [-4.8,+1.5] \\
GPT-6 Astra & All pages & low & 128 & 45 & -0.8 [-2.2,+0.0] \\
GPT-6 Astra & CB-high EM miss & high & 100 & 39 & -2.0 [-6.4,+0.6] \\
GPT-6 Astra & CB-high EM miss & low & 100 & 39 & -1.0 [-2.7,+0.0] \\
GPT-6 Astra & CB-high EM hit & high & 28 & 18 & +1.8 [+0.0,+6.2] \\
GPT-6 Astra & CB-high EM hit & low & 28 & 18 & +0.0 [+0.0,+0.0] \\
GPT-6 Astra & Both CB: no later alias & high & 92 & 37 & -2.2 [-7.1,+0.6] \\
GPT-6 Astra & Both CB: no later alias & low & 92 & 37 & -1.1 [-3.0,+0.0] \\
Claude Fable 5.1 & All pages & high & 128 & 45 & -0.4 [-1.9,+0.9] \\
Claude Fable 5.1 & All pages & low & 128 & 45 & -2.3 [-6.3,+1.1] \\
Claude Fable 5.1 & CB-high EM miss & high & 118 & 44 & -0.4 [-2.0,+1.0] \\
Claude Fable 5.1 & CB-high EM miss & low & 118 & 44 & -2.5 [-6.4,+0.9] \\
Claude Fable 5.1 & CB-high EM hit & high & 10 & 7 & +0.0 [+0.0,+0.0]\,$^{p}$ \\
Claude Fable 5.1 & CB-high EM hit & low & 10 & 7 & +0.0 [-15.0,+15.0]\,$^{p}$ \\
Claude Fable 5.1 & Both CB: no later alias & high & 96 & 39 & -1.0 [-2.9,+0.0] \\
Claude Fable 5.1 & Both CB: no later alias & low & 96 & 39 & -2.6 [-7.5,+1.5] \\
\bottomrule
\end{tabular}
\caption{All frozen API subsets, including descriptive sensitivities, without outcome-based selection. Miss/hit use the complete high-effort closed-book EM result; both-CB-no-later-alias requires complete responses without a later alias at both efforts. Nominal 95\% predicate intervals are conditional on subset membership; $p$ marks a page-only interval when fewer than ten predicate groups remain. These lexical partitions do not certify knowledge or its absence.}
\label{tab:n3-subsets}
\end{table*}

\section{Synthetic Records and Question-Date Spelling}
\label{app:m12}

\paragraph{Fixed construction.}
Before generating responses, we fixed 128 fictional records: four relations
(headquarters, coach, maintainer, operator), two styles (one-sentence prose
or key--value), and 16 records per relation/style. Deterministic nonce
entity/answer names are disjoint. Each newer revision replaces the older
state; both precede the later question. Dates vary across 2024--2026.
The original system instruction is retained. Each record has both
$(0,+1)/(+1,0)$ assignments in both orders, a two-version no-collection-field
control in both orders, and one newer-only no-field control. These last
controls retain the collection definition; they are not the earlier
definition-shortened omission condition. No record is filtered by answers.
Record-bootstrap intervals condition on the eight fixed templates; they
do not estimate generalization to a population of templates or natural QA.
\begin{table*}[t]
\centering\footnotesize\setlength{\tabcolsep}{2pt}
\begin{tabular}{llllll}
\toprule
LLM & Newer only & No field & Newer matches & Older matches & Swap: earlier EM \\
\midrule
Qwen2.5-7B & 99.2 / 0.0 & 97.3 / 2.3 & 98.4 / 0.0 & 2.7 / 96.1 & \shortstack{+96.1\\{[+93.8,+98.0]}} \\
Phi-4 & 93.0 / 0.0 & 96.9 / 0.0 & 94.5 / 0.0 & 0.0 / 94.5 & \shortstack{+94.5\\{[+90.6,+97.7]}} \\
Mistral-24B & 99.2 / 0.0 & 99.6 / 0.0 & 99.6 / 0.4 & 5.5 / 94.5 & \shortstack{+94.1\\{[+91.4,+96.9]}} \\
Qwen3.5-35B & 100.0 / 0.0 & 100.0 / 0.0 & 100.0 / 0.0 & 7.4 / 92.6 & \shortstack{+92.6\\{[+89.5,+95.3]}} \\
Qwen2.5-72B & 100.0 / 0.0 & 100.0 / 0.0 & 99.6 / 0.0 & 7.4 / 91.4 & \shortstack{+91.4\\{[+87.9,+94.5]}} \\
\bottomrule\end{tabular}
\caption{128 fictional records, both revision orders except the single newer-only input. No-field controls retain the original collection definition. Levels give later-reference ($t_1$) / earlier-reference ($t_0$) EM (\%). The final column gives the earlier-reference swap change, with a nominal 95\% record interval conditional on eight templates.}
\label{tab:m1-levels}
\end{table*}

\begin{table*}[t]
\centering\footnotesize\setlength{\tabcolsep}{2pt}
\begin{tabular}{lll}
\toprule
LLM & Prose (64 records) & Key--value (64 records) \\
\midrule
Qwen2.5-7B & -96.1 [-99.2,-92.2] & -95.3 [-98.4,-91.4] \\
Phi-4 & -90.6 [-96.1,-84.4] & -98.4 [-100.0,-95.3] \\
Mistral-24B & -93.8 [-97.7,-89.1] & -94.5 [-97.7,-90.6] \\
Qwen3.5-35B & -96.9 [-99.2,-93.8] & -88.3 [-93.0,-82.8] \\
Qwen2.5-72B & -94.5 [-98.4,-90.6] & -89.8 [-94.5,-85.2] \\
\bottomrule\end{tabular}
\caption{Synthetic question-date matching swap effects on later-reference EM, in both render styles. Each style has 16 records per relation. Nominal record-bootstrap intervals condition on four relation templates; style results are descriptive, with no selected-style significance claims.}
\label{tab:m1-styles}
\end{table*}

\begin{table*}[t]
\centering\footnotesize\setlength{\tabcolsep}{2pt}
\begin{tabular}{llllll}
\toprule
LLM & \shortstack{ISO: newer\\matches} & \shortstack{ISO: older\\matches} & \shortstack{English: newer\\matches} & \shortstack{English: older\\matches} & \shortstack{English:\\earlier EM} \\
\midrule
Qwen2.5-7B & 64.5 / 7.0 & 27.0 / 42.2 & 66.0 / 5.5 & 31.2 / 40.2 & \shortstack{+34.8\\{[+24.6,+44.7]}} \\
Phi-4 & 49.6 / 2.3 & 6.6 / 42.6 & 46.9 / 2.3 & 15.2 / 31.6 & \shortstack{+29.3\\{[+19.2,+39.7]}} \\
Mistral-24B & 63.7 / 4.7 & 16.0 / 50.0 & 69.1 / 2.7 & 16.0 / 53.1 & \shortstack{+50.4\\{[+37.1,+63.0]}} \\
Qwen3.5-35B & 68.0 / 5.9 & 28.5 / 45.7 & 68.4 / 5.9 & 32.4 / 41.0 & \shortstack{+35.2\\{[+24.6,+45.5]}} \\
Qwen2.5-72B & 73.4 / 1.2 & 20.3 / 48.8 & 75.0 / 0.4 & 28.5 / 41.8 & \shortstack{+41.4\\{[+29.3,+55.4]}} \\
\bottomrule\end{tabular}
\caption{128 Wikipedia pages, 45 predicate groups, both orders and all four spelling/assignment conditions. Levels give later-reference ($t_1$) / earlier-reference ($t_0$) EM (\%). The final column gives the English-format earlier-reference swap change with a nominal 95\% predicate-cluster interval. ISO controls are rerun alongside the English-date conditions.}
\label{tab:m2-levels}
\end{table*}

\begin{table*}[t]
\centering\footnotesize\setlength{\tabcolsep}{3pt}
\begin{tabular}{lllllll}
\toprule
API LLM & Effort & Newer matches & Older matches & $\Delta$ later EM & $\Delta$ earlier EM & Later alias \\
\midrule
GPT-6 Astra & high & 100.0 & 100.0 & +0.0 [+0.0,+0.0] & +0.0 [+0.0,+0.0] & 100.0 \\
GPT-6 Astra & low & 100.0 & 99.6 & -0.4 [-1.2,+0.0] & +0.0 [+0.0,+0.0] & 99.6 \\
Claude Fable 5.1 & high & 100.0 & 100.0 & +0.0 [+0.0,+0.0] & +0.0 [+0.0,+0.0] & 100.0 \\
Claude Fable 5.1 & low & 100.0 & 85.9 & -14.1 [-18.0,-10.2] & +0.4 [+0.0,+1.2] & 99.6 \\
\bottomrule\end{tabular}
\caption{Question-date matching swaps on 128 fictional records, 2,048 API responses. Newer/older matching conditions give later-reference ($t_1$) EM (\%) when the newer/older revision receives $t_1$. Changes are older-matching minus newer-matching percentage points; later/earlier EM uses $t_1$/$t_0$ references. Later alias is later-answer inclusion in older-matching responses, not necessarily unambiguous selection. Nominal 95\% paired-record intervals condition on eight templates, averaging both orders; [0,0] means no observed variation, not equivalence. Both efforts use reasoning.}
\label{tab:synthetic-api}
\end{table*}

\paragraph{Spelling intervention.}
The previously exposed 128-page disjoint rerun sample has both assignments
and orders with either the original ISO question date or its English
month-name equivalent. Only the final Requested time value changes;
question text, collection/revision ISO fields and evidence bytes remain
fixed. ISO controls are rerun alongside the English-date conditions. The plan fixes
ISO/English swaps and their interaction, not a subset selected by outcome.
All five LLMs use the original greedy settings; all 9,600 responses
complete without caps, empty finals or retries.
The five synthetic and five English-format EM swap effects survive the
Holm correction across all 60 follow-up tests. Of the five spelling interactions,
only Phi-4's attenuation survives; all other interactions remain unresolved.

\paragraph{Synthetic API follow-up.}
\label{app:synthetic-api}
The same 128 invented records receive both assignments and source orders
at high/low effort in each API LLM: 1,024 calls per LLM, 2,048 total
(Table~\ref{tab:synthetic-api}). Message contents are unchanged from the
local experiment, with its system instruction moved verbatim to the
single-user prefix as in the Wikipedia-revision API study. Settings and independent
requests are otherwise as above; no additional no-field control is run.
The fixed plan precedes these outputs. Record-paired inference conditions
on the same eight templates. The six EM tests include both effort effects
and their interaction per LLM; 12 exclusive-alias tests give an 18-test
union, with a 78-test sensitivity including the local 60. All levels,
contrasts and style/relation strata remain in the numerical archive.
All 2,048 finals complete, without retries or caps. Claude low has 36
nonexact older-matching responses: 35 retain the later alias in longer answers,
including 21 with both aliases; one contains only the earlier alias. No
post hoc extraction replaces the frozen whole-response EM score.

\section{Selection and Reproduction}
\label{app:selection}

Selection retains the latest supplied revision timestamped by the question
date. Table~\ref{tab:earlier-selection} uses paired controls from the same
595-page plain-display execution, not the main grid or 596-page replication.
The 15.0--30.9-point earlier-question gains concern revision recovery;
selection can discard later corrections. Input-token accounting in the
technical supplement is not a latency or energy measurement.

The reproduction package covers ten principal experiment blocks:
the two assignment grids, past-side and prompt controls, synthetic and
spelling tests, expanded thinking, native reasoning, and both API studies.
It contains exact messages, reference answers, generated outputs and
offline scoring code, with 160,484 responses and 5,060 numerical checks.
Historical shared-date, selection and external-data diagnostics remain
separately documented in the technical results. Code, prompts and outputs
for the principal experiments will be publicly released on acceptance.

\end{document}